\pdfoutput=1
\documentclass[10pt,twocolumn,letterpaper]{article}

\usepackage{iccv}              % To produce the CAMERA-READY version
\definecolor{iccvblue}{rgb}{0.21,0.49,0.74}
\usepackage[pagebackref,breaklinks,colorlinks,allcolors=iccvblue]{hyperref}
\usepackage{multirow}
\usepackage{booktabs}
\usepackage{tabularx}
\usepackage{ragged2e}
\usepackage{array}
\usepackage{longtable}
\usepackage{float}
\usepackage{stfloats}
\usepackage[accsupp]{axessibility}

\newcolumntype{Y}{>{\centering\arraybackslash}X}
\newcolumntype{M}[1]{>{\centering\arraybackslash}m{#1}} % 核心：这个 M 列类型才能让 Method 单元格垂直居中
\def\paperID{*****} % *** Enter the Paper ID here
\def\confName{ICCV}
\def\confYear{2025}

\title{Benchmarking Out-of-Distribution Detection for Plankton Recognition: A Systematic Evaluation of Advanced Methods in Marine Ecological Monitoring}

\author{Yingzi Han$^{\ast}$\\
Beijing Normal University\\
China\\
{\tt\small hanyingzi@mail.bnu.edu.cn}
\and
Jiakai He$^\ast$\\
Beijing Normal University\\
 China\\
{\tt\small hejiakai@mail.bnu.edu.cn}
\and 
Chuanlong Xie$^{\dagger}$\\
Beijing Normal University\\
 China\\
{\tt\small clxie@bnu.edu.cn}
\and 
Jianping Li\\
Shenzhen Institutes of Advanced Technology, Chinese Academy of
Sciences\\
China\\
{\tt\small jp.li@siat.ac.cn}
}

\begin{document}
\maketitle
\begin{abstract}
% 尽管深度学习在自动化浮游生物识别方面取得了显著进展，然而，当模型部署到实际海洋环境中时，由于浮游生物形态的复杂性、巨大的物种多样性、环境噪声以及持续发现的新物种，模型不可避免地会遭遇训练与测试数据分布不一致（即分布外，OoD）的情况，这导致在推理和测试阶段出现不可预测的错误。尽管近年来OoD检测方法取得了重大突破，但在浮游生物识别任务中，相关研究仍缺乏对计算机视觉领域最新进展的系统整合，并且缺乏一个统一的基准进行大规模评估。

% 本文基于DYB-PlanktonNet数据集，精心设计了一系列模拟不同分布变化场景的OoD基准，旨在弥补这一空白。我们系统地评估了二十二种目前最先进的OoD检测方法，并利用定量指标对其性能进行了全面的评估与比较。在所设计的不同测试基准下，我们进行了大量的实验，以充分评估这些OoD检测方法的整体性能。比较结果表明，在我们构建的基准中，ViM方法在综合性能上显著优于其他类型的方法，尤其是在远分布外（Far-OoD）情景中表现出色。

% 这项全面的评估不仅为自动化浮游生物识别领域的算法选择提供了可靠的参考，更为未来浮游生物OoD检测研究奠定了坚实的基础。据我们所知，本研究是首次对浮游生物识别中分布外数据检测方法进行大规模、系统性的评估和分析。
Automated plankton recognition models face significant challenges during real-world deployment due to distribution shifts (Out-of-Distribution, OoD) between training and test data. This stems from plankton's complex morphologies, vast species diversity, and the continuous discovery of novel species, which leads to unpredictable errors during inference. Despite rapid advancements in OoD detection methods in recent years, the field of plankton recognition still lacks a systematic integration of the latest computer vision developments and a unified benchmark for large-scale evaluation. To address this, this paper meticulously designed a series of OoD benchmarks simulating various distribution shift scenarios based on the DYB-PlanktonNet dataset \cite{875n-f104-21}, and systematically evaluated twenty-two OoD detection methods. Extensive experimental results demonstrate that the ViM \cite{wang2022vim} method significantly outperforms other approaches in our constructed benchmarks, particularly excelling in Far-OoD scenarios with substantial improvements in key metrics. This comprehensive evaluation not only provides a reliable reference for algorithm selection in automated plankton recognition but also lays a solid foundation for future research in plankton OoD detection. To our knowledge, this study marks the first large-scale, systematic evaluation and analysis of Out-of-Distribution data detection methods in plankton recognition. Code is available at \url{https://github.com/BlackJack0083/PlanktonOoD}. 
\end{abstract}    
\footnotetext[1]{$\ast$ Equal contribution.}
\footnotetext[2]{$\dagger$ Corresponding author.}
\section{Introduction}
\label{sec1:intro}

Plankton constitutes a fundamental component of marine ecosystems, playing a pivotal role in maintaining ecological balance, participating in global carbon cycles, and supporting marine food webs. The species composition, abundance, and distribution dynamics of plankton not only directly impact normal human life and production activities but also play a critical role in assessing marine environmental health and research on climate change early warning systems \cite{murphyCascadingEffectsClimate2020}. In recent years, with the widespread adoption of underwater imaging devices and the rapid development of deep learning techniques, automated plankton recognition has emerged as one of the core approaches in marine ecological monitoring \cite{pastore2020annotationfree, pu2021anomaly, ciranniComputerVisionDeep2024}. However, the morphological complexity and immense species diversity of plankton pose significant challenges for automatic classification systems, as inter-species differences are often subtle and difficult to discern \cite{eerola2024survey, kareinen2025open}. In addition, automatically acquired plankton images frequently contain substantial amounts of noise from non-plankton organisms, as well as potential instances of previously undiscovered or unannotated species. These factors necessitate that any pretrained plankton recognition model deployed in real-world marine environments must possess the capability to distinguish between known and unknown categories.

Current mainstream approaches generally treat plankton image recognition as a K+1 classification problem, with K referring to the specific plankton categories of interest and the extra class representing the non-target background \cite{yang2022contrastive, walker2021improving}. The earliest studies in planktonic organism image classification primarily relied on handcrafted features. This approach necessitated extensive expert knowledge, offered strong interpretability, and provided striking ecological and biogeochemical insights \cite{blaschko2005automatic, sosik2007automated}.

However, treating this task as a conventional K+1 classification problem requires the training data to contain sufficiently representative samples of the ``1'' background class. In practice, however, this background class is open-ended and highly diverse, making this assumption difficult to satisfy in real-world scenarios. Therefore, the problem of recognizing whether a sample belongs to this background class is sometimes reformulated as a one-sample hypothesis testing problem, where the goal is to determine whether a given test image does not belong to any of the K known classes, based solely on the observations from these K classes \cite{xue2024enhancing}. 

With the development of deep learning, a common solution is to use deep neural networks to automatically extract image features, which are then employed for score-based decision making to determine whether a given sample belongs to the known distribution. Such an approach is referred to as Out-of-Distribution (OoD) detection. In this paradigm, a post hoc classifier assigns a confidence or similarity score to the feature representation, which is then compared against a predefined threshold to determine whether the sample is In-Distribution (ID) or OoD. Pu \etal~\cite{pu2021anomaly} explored the use of the Mahalanobis Distance for OoD detection and suggested that Maximum Softmax Probability (MSP) and energy-based methods are also promising directions. Yang \etal~\cite{yang2022contrastive} trained a feature extractor using supervised contrastive learning to obtain more discriminative representations and employed cosine similarity as the metric. Similarly, Ciranni \etal~\cite{ciranni2024anomaly} applied Principal Component Analysis (PCA) to the features and trained a separate one-class SVM for each known class; samples are detected as OoD if they fail to meet the threshold criteria across all classifiers. Collectively, these studies offer initial empirical support for the effectiveness of integrating neural network feature extraction with post hoc strategies for reliable OoD detection.

Although the aforementioned studies have paid considerable attention to the openness and complexity of the plankton background class and have adopted dedicated OoD detection methods to address this issue, their design and application of scoring functions remain relatively naive, often relying on conventional approaches such as MSP, Mahalanobis Distance, or inner product similarity. Despite the substantial advances in OoD detection methods since 2017, the diversity of scoring functions has not been fully exploited in existing work in the field of plankton detection, even though it holds great potential for improving the recognition of the ``1'' (background) class.

 Extensive prior research indicates that the performance of different post hoc classifiers varies depending on the dataset and task, and that no single post hoc technique consistently outperforms others in all scenarios \cite{DBLP:journals/corr/abs-1809-04729, rs16091501}. Techapanurak and Okatani \cite{techapanurak2021practical} compared several OoD scores across multiple datasets and found that the Mahalanobis method performs well only for detecting inputs far from the training distribution, and the discriminative performance of MCDropout on domain shift caused by image corruption improves dramatically with stronger pre-training. Tajwar \etal~\cite{tajwar2021no} found that distance-based OoD detection methods are easily confused by ID samples that lie close to the detection boundary, leading to a rapid drop in performance. Moreover, the effectiveness of different scores varies to different extents depending on the amount of available ID data. Therefore, for the specific needs in plankton detection, it's essential to establish a comprehensive evaluation framework covering mainstream OoD detection methods, which would allow for the practical selection of suitable detection methods for real-world ecological monitoring tasks.

Furthermore, existing studies often rely on datasets that differ significantly from the ID imaging conditions when constructing OoD benchmarks \cite{pu2021anomaly, yang2022contrastive}. This may cause models to exploit spurious correlations rather than learning essential discriminative features. Furthermore, lumping all OoD samples into a singular ``unknown class'' fails to adequately assess a model's proficiency in detecting various types of open data during real-world deployment. To address these challenges, we partitioned the dataset collected from Daya Bay, Shenzhen, into three parts: the In-Distribution (ID) subset containing ecologically significant species (\eg \textit{Jellyfish} and \textit{Creseis acicula}, whose abnormal proliferation may signal environmental change and potentially clog nuclear power plant outlets \cite{zhao2022role, wang2023aggregation, zhang2025source, zeng2021acoustic}), the Near-OoD subset consisting of less ecologically significant plankton species, and the Far-OoD subset comprising noise images such as fish eggs and bubbles. We evaluated twenty-two OoD detection methods on our established benchmark and conducted a comprehensive analysis of the experimental results. 

The main contributions of this work are summarized as follows:
\begin{itemize}
  \item We established a systematic OoD detection benchmark for plankton recognition. 
  \item We conducted a comprehensive evaluation of various mainstream OoD post hoc methods, providing a reliable reference for algorithm selection in the field of automated plankton recognition.
  \item We analyzed the performance discrepancies and challenges of these OoD detection methods when applied to the real-world classification of plankton.
\end{itemize}

\section{Preliminaries}
\label{sec2:related}

\subsection{Plankton Background Class Detection}

Background class detection is a critical problem in underwater ecological vision \cite{wyatt2025safe, nawaz2025survey, saleh2024applications}. In the context of plankton analysis, in addition to framing it as an out-of-distribution (OoD) detection task as explained in \cref{Out-of-Distribution Detection}, previous studies have often approached it as an anomaly detection or open-set recognition problem, highlighting how different problem assumptions can lead to distinct solution strategies.

Anomaly detection refers to the problem of finding patterns in data that do not conform to expected behavior \cite{chandola2009anomaly}. Varma \etal~\cite{varma2020autonomous} proposed an anomaly detection method based on L1-norm tensor conformity to eliminate misclassified or non-plankton samples from the training dataset by evaluating their consistency in low-rank subspaces~\cite{tountas2019conformity}. Pastore \etal~\cite{pastore2020annotationfree} trained a DEC detector for each training species, specifically one for each plankton species identified in the unsupervised learning step, achieving superior performance compared to the one-class SVM.

Open set recognition (OSR) assumes that recognition in the real world is an open-set problem, meaning that the recognition system should reject unknown or unseen classes at test time. A common approach to achieve this is to formulate it as a similarity metric learning problem. Teigen \etal~\cite{teigen2020leveraging} employed a Siamese network trained with triplet loss to evaluate few-shot learning and novel class detection scenarios. Badreldeen \etal~\cite{mohamed2022deep} further adopted angular margin loss (ArcFace)~\cite{deng2019arcface} in place of triplet loss and utilized generalized mean pooling (GeM)~\cite{radenovic2018finetuning} to produce rotation- and translation-invariant features.

\subsection{Out-of-Distribution Detection}
\label{Out-of-Distribution Detection}

Out-of-Distribution (OoD) detection refers to the task of determining whether a test input is drawn from the same data distribution as the training set. Formally, let $\mathcal{X}$ and $\mathcal{Y}$ denote the input and label spaces, respectively, and let $P_0$ represent the joint distribution over $\mathcal{X} \times \mathcal{Y}$ for the training data. The marginal distribution of inputs is denoted by $P_X$. A sample $x \sim P_X$ is referred to as an In-Distribution (ID) example, whereas a sample drawn from an unknown distribution $Q$ ($Q \neq P_X$) is considered as an OoD sample.

The OoD detection task can be naturally formulated as a statistical hypothesis testing problem:
$$
H_0: x^\ast \sim P_X \quad \text{vs.} \quad H_1: x^\ast \sim Q, \quad Q \in \mathcal{Q}, \, P_X \notin \mathcal{Q}
$$
where $x^\ast$ denotes a test input, and $\mathcal{Q}$ represents a family of possible OoD distributions.

% In standard classification tasks, we typically assume test data to be independent and identically distributed (i.i.d.) with respect to the training data distribution. However, OoD detection violates this fundamental assumption, necessitating specialized methodological approaches. 

In practice, OoD detection is typically implemented with a score function $S(x; \phi)$, where $\phi$ denotes a neural network feature extractor or classifier, and $S(\cdot; \phi)$ assigns higher scores to ID samples and lower scores to OoD samples. A decision rule is applied as:
\begin{equation}
    G(x^\ast; \phi) = 
    \begin{cases}
    \text{ID}, & \text{if } S(x^\ast; \phi) > \lambda_\phi, \\
    \text{OoD}, & \text{if } S(x^\ast; \phi) \le \lambda_\phi
    \end{cases}
\end{equation}
where $\lambda_\phi$ is a predefined threshold controlling the trade-off between true positive rate and false positive rate.

It's worth noting that when we change the null hypothesis, meaning we select a different class as the positive class to calculate the false positive rate (FPR) at a given true positive rate (TPR), the results can differ significantly. As demonstrated in  \cref{tab:far_ood_performance} and \cref{tab:near-OoD results}, the false positive rates exhibit significant divergence depending on whether In-Distribution (ID) or Out-of-Distribution (OoD) samples are designated as the positive class. However, in real-world applications, valuable plankton images are rare and precious, while noise images constitute the vast majority. Therefore, the majority of existing works adopt ID samples as the positive class. 

Recent advances in OoD detection have led to a wide range of post-hoc methods, which are categorized in \cref{tab:method_categories}. In this study, we systematically evaluated mainstream OoD detection methods proposed over the years on our plankton datasets. While these techniques have demonstrated excellent performance on general computer vision benchmarks, their robustness and generalizability remain limited when confronted with the challenges posed by plankton images, such as complex backgrounds, substantial intra-class diversity, and the frequent presence of unknown species.

\begin{table*}[!ht]
    \setlength{\belowcaptionskip}{-0.4cm}
    \centering
    \begin{tabular}{>{\centering\arraybackslash}p{0.2\textwidth}>{\centering\arraybackslash}p{0.5\textwidth}>{\centering\arraybackslash}p{0.2\textwidth}}
    \hline
        \textbf{Distance-based} & \textbf{Classification-based} & \textbf{Density-based} \\ \hline
        Mahalanobis \cite{lee2018simple} & ViM \cite{wang2022vim},  Residual \cite{zisselman2020deep}, ODIN \cite{liang2018enhancing},  GEN \cite{liu2023gen}, MSP \cite{hendrycks2017baseline} & Energy \cite{liu2020energy} \\ 
        RMDS \cite{ren2021simple}, KNN \cite{sun2022out} & OpenMax  \cite{bendale2016towards}, Relation \cite{kim2023neural},  TempScale \cite{guo2017calibration}, & DICE \cite{sun2022dice}\\ 
        fDBD \cite{liu2023fast} & MCDropout \cite{gal2016dropout}, KL Matching  \cite{basart2022scaling}, GradNorm \cite{huang2021importance} & ~ \\ 
        ~ & MLS \cite{basart2022scaling}, ReAct  \cite{sun2021react}, ASH \cite{djurisic2022extremely}, SHE \cite{zhang2022out}, RankFeat \cite{song2022rankfeat} & ~ \\ 
    \hline
    \end{tabular}
    \caption{Post Hoc Methods for OoD Detection. For a detailed description of each method, please refer to the Appendix \ref{oodmethods_intro}.}
    \label{tab:method_categories}
\end{table*}

\section{Dataset Construction and Analysis}
\label{sec3: dataset}

Our dataset is derived from DYB-PlanktonNet \cite{875n-f104-21}, a publicly available dataset of marine plankton and suspended particles from Daya Bay. Motivated by practical marine ecological monitoring needs, we adopt a methodology from \cite{kim2023unified, zhang2023openood, wang2025dissecting} to partition the 92 original categories into distinct In-Distribution (ID) and various Out-of-Distribution (OoD) subsets. This stratified partitioning is inspired by generalized OoD detection \cite{yang2024generalized}, which expands beyond the traditional domain-disjoint definition. Our approach addresses three key challenges: in-domain semantic shifts (Near-OoD), in-domain non-biological clutter (Far-OoD (Bubbles \& Particles)), and out-of-domain shifts represented by external datasets (Far-OoD (General)). This fine-grained categorization enables a more precise and realistic evaluation of OoD detection performance than prior work that treated all non-target entities as a single background class. The detailed data category division is as follows:

\begin{figure*}[h]
  \centering
  \includegraphics[width=1\linewidth]{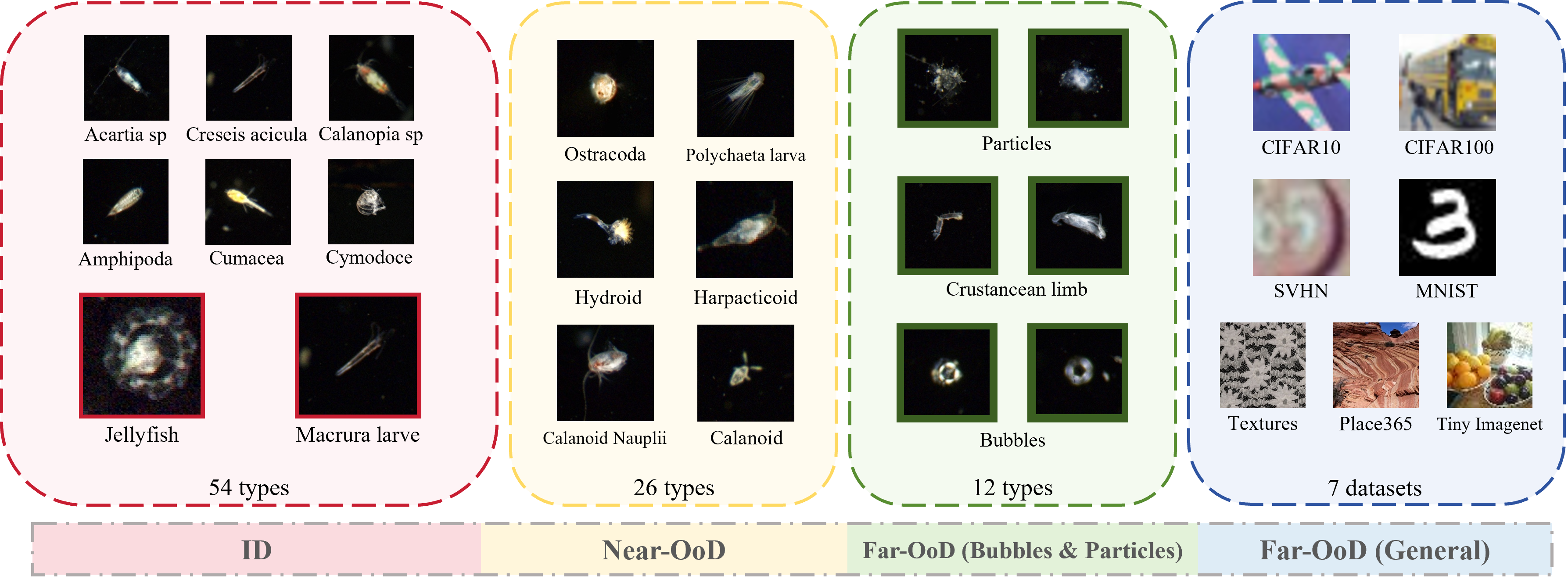}
  \caption{Our constructed plankton Out-of-Distribution detection image benchmark comprises four distinct distribution shift scenarios: ID, Near-OoD, Far-OoD (Bubbles \& Particles), and Far-OoD (General). For each distribution, we provide representative class images. A detailed classification can be found in the Supplementary Material.}
  \label{fig:datasets}
\end{figure*}

\noindent
\textbf{ID data:} We define 54 categories as In-Distribution (ID) data, comprising abundant samples of native or parasitic plankton commonly observed in Daya Bay water intake. These include ecologically significant groups like \textit{Jellyfish} (potential cooling system cloggers) and \textit{Creseis acicula} (linked to abnormal blooms) \cite{zhao2022role, wang2023aggregation, zhang2025source, zeng2021acoustic}. These categories serve as primary detection targets for routine monitoring and constitute the ID class space for model training and evaluation.

\noindent
\textbf{Near-OoD data:} This subset comprises 26 biological categories that are morphologically or ecologically related to the ID classes but exhibit lower sample frequency or less direct monitoring importance. It includes larval stages of certain plankton and uncommon forms such as \textit{Hydroid} (gelatinous zooplankton) and \textit{Ostracoda} (small crustaceans). These examples represent semantically similar yet non-core taxa, and are used to define the Near-OoD subset, simulating ``novel-but-similar" plankton species that a deployed model might encounter.

\noindent
\textbf{Far-OoD (Bubbles \& Particles) data:} We further designate 12 categories as Far-OoD examples that exhibit significant semantic deviation from known plankton class. These are primarily non-biological entities or artifacts introduced during image acquisition, such as bubbles, body fragments, and environmental particles. While they bear little ecological relevance, their presence in raw image streams poses practical challenges for robust OoD detection. This subset aims to model real-world imaging noise and clutter frequently encountered in plankton monitoring systems. Notably, these Far-OoD (Bubbles \& Particles) categories, alongside the Near-OoD categories, collectively constitute the background class within our benchmark. These represent non-target entities that a deployed model must identify and differentiate in real-world scenarios. 

\noindent
\textbf{Far-OoD (General) data:} To comprehensively assess the robustness and generalization ability of OoD methods, we incorporate additional benchmark datasets widely adopted in the computer vision community. These include \texttt{CIFAR-10} \cite{krizhevsky2009learning}, \texttt{CIFAR-100} \cite{krizhevsky2009learning}, \texttt{SVHN} \cite{netzer2011reading}, \texttt{Texture} \cite{cimpoi2014describing},  \texttt{MNIST} \cite{deng2012mnist}, \texttt{Places365} \cite{zhou2017places}, and \texttt{Tiny ImageNet} \cite{torralba200880}.  These datasets contain objects and scenes semantically unrelated to the marine domain, serving as strong Far-OoD samples that do not naturally occur in plankton imagery. We refer to this group as the Far-OoD (General) subset, representing disjoint visual domains.

In total, we construct four well-defined subsets: ID, Near-OoD, Far-OoD (Bubbles \& Particles), and Far-OoD (General), as shown in \cref{fig:datasets}. This stratified partitioning provides a realistic and challenging benchmark for OoD detection in marine plankton scenarios. The complete category lists for each subset are provided in the Appendix \ref{sec:dataset}.
\section{Experiments}
\label{sec4:experiment}

This section details our systematic evaluation of methods on the plankton OoD detection benchmark constructed in \cref{sec3: dataset}. We evaluate the performance of all post hoc OoD detection methods mentioned in \cref{sec2:related}, specifically on both Far-OoD and Near-OoD benchmark, strictly adhering to the OpenOOD-v1.5 \cite{zhang2023openood} evaluation protocol. For performance evaluation, we employ the widely recognized metrics of FPR95 and AUROC, further incorporating the more stringent FPR99 to provide comprehensive performance.

\subsection{Experimental Settings}

\noindent
\textbf{Experiments Metrics.} To comprehensively evaluate the performance of OoD methods, we adopt a set of widely accepted metrics to ensure both robustness and fairness in the assessment. These metrics are commonly used in the existing OoD detection literature. Considering the inherent class imbalance in real-world marine plankton datasets, we report results from two complementary perspectives: one treating In-Distribution (ID) samples as the positive class, and the other treating Out-of-Distribution (OoD) samples as the positive class. The latter approach follows the evaluation protocol introduced by OpenOOD-v1.5 \cite{zhang2023openood}, offering a more complete view of detector performance. The main evaluation metrics are as follows:

\begin{itemize}
    \item \textbf{False Positive Rate at 95\% and 99\% TPR on ID samples (FPR\@95-ID, FPR\@99-ID):} These metrics quantify the proportion of OoD samples misclassified as ID when ID detection achieves 95\% and 99\% true positive rates (TPR). This aligns with our marine plankton monitoring goal: high recall for key species while filtering irrelevant OoD instances.
    \item \textbf{False Positive Rate at 95\% and 99\% TPR on OoD samples (FPR\@95-OoD, FPR\@99-OoD):} Conversely, these metrics evaluate the proportion of ID samples mistakenly identified as OoD when OoD detection reaches 95\% and 99\% TPR. This matches standards from large-scale OoD benchmarks like OpenOOD-v1.5 \cite{zhang2023openood}, enabling fair comparisons.
    \item \textbf{Area Under the Receiver Operating Characteristic Curve (AUROC):} AUROC quantifies the detector's overall discriminative ability, representing the probability that a randomly selected positive sample ranks higher than a negative one. It offers a threshold-independent performance measure across all decision boundaries.
    \item \textbf{ID classification accuracy (ACC):} Reflects the network's classification accuracy on In-Distribution (ID) samples, indicating its ability to correctly recognize known categories.
\end{itemize}

\noindent
\textbf{Remark on the Implementation.} All experiments are implemented using PyTorch 2.4.1. Our evaluation framework is built upon OpenOOD-v1.5 \cite{zhang2023openood}, a comprehensive benchmarking platform for Out-of-Distribution detection. We rigorously test twenty-two post hoc OoD detection methods provided mentioned in \cref{tab:method_categories}. These methods can be broadly categorized according to their underlying principles into:
(1) classification-based approaches,
(2) density-based approaches, and
(3) distance-based approaches.
This systematic evaluation aims to explore and demonstrate the applicability and potential of modern OoD detection techniques in the context of marine science.

\noindent
\textbf{Network Architectures and Training Protocol.}  To ensure a comprehensive evaluation of OoD detection performance across different network architectures, we constructed a diverse model zoo comprising both popular and robust deep neural architectures. This includes ResNet-18, ResNet-50, ResNet-101, ResNet-152 \cite{he2016deep}, DenseNet-121, DenseNet-169, DenseNet-201 \cite{huang2017densely}, SE-ResNeXt-50 \cite{hu2018squeeze} and ViT \cite{dosovitskiy2020image}. ResNet \cite{he2016deep} introduces residual connections to address the vanishing gradient and model degradation issues in deep network training, allowing for effective training of very deep networks and improving performance. DenseNet \cite{huang2017densely} maximizes information flow, promotes feature reuse, and reduces parameters through dense inter-layer connections. SE-ResNeXt \cite{hu2018squeeze} combines the Squeeze-and-Excitation module \cite{hu2018squeeze} with the ResNeXt  \cite{xie2017aggregated} architecture, where the former enhances representational power by learning channel attention, and the latter improves efficiency and accuracy through grouped convolutions. ViT \cite{dosovitskiy2020image} applies a standard Transformer encoder to image patches, treating image classification as a sequence-to-sequence prediction. It achieves strong performance by leveraging self-attention.
 These architectures are widely adopted in the OoD detection literature and offer a varied set of feature extractors. \Cref{tab:classifier} summarizes the specifications of the above architectures. All backbone models were trained from scratch on the ID dataset's training split, using softmax cross-entropy (CE) loss. We trained each model for 100 epochs using stochastic gradient descent (SGD) with a momentum of 0.9. The initial learning rate was set to 0.1 and adjusted using a cosine annealing schedule. A weight decay of 5×10-4 was applied to regularize the training. For each network architecture, we repeated the training three times using different random seeds to ensure robustness. For each post hoc OoD detection method, we report the best performance achieved across all backbones in our model zoo. In other words, the final results for each OoD method are based on its most compatible and highest-performing backbone model. 

\begin{table}[!htbp]
    \centering
    \begin{tabularx}{\linewidth}{
        p{3cm}  % 第一列宽度 3cm
        >{\centering\arraybackslash}X  % 第二列居中+自动填充
        >{\centering\arraybackslash}X  % 第三列居中+自动填充
    }
    \hline
    \textbf{Classifier} & \textbf{Params} & \textbf{ACC(\%)} \\ \hline
    ResNet-18 \cite{he2016deep} & 11.69M & 95.42\scriptsize $\pm$0.24 \\ 
    ResNet-50 \cite{he2016deep} & 25.56M & 94.92\scriptsize $\pm$0.15 \\ 
    ResNet-101 \cite{he2016deep} & 44.55M & 95.06\scriptsize $\pm$0.29 \\ 
    ResNet-152 \cite{he2016deep} & 60.19M & 95.00\scriptsize $\pm$0.34 \\ 
    DenseNet-121 \cite{huang2017densely} & 7.98M & 96.15\scriptsize $\pm$0.20 \\ 
    DenseNet-169 \cite{huang2017densely} & 14.14M & 95.94\scriptsize $\pm$0.16 \\ 
    DenseNet-201 \cite{huang2017densely} & 20.01M & 96.06\scriptsize $\pm$0.13 \\ 
    SE-ResNeXt-50 \cite{hu2018squeeze} & 28.07M & 95.65\scriptsize $\pm$0.30 \\ 
    ViT \cite{dosovitskiy2020image} & 86.57M & 90.49\scriptsize $\pm$0.15\\ \hline
    \end{tabularx}
    \caption{Specifications of different architectures: the number of parameters and ID classification accuracy (ACC) on the ID data testing subset. All ACC values are reported as the mean $\pm$ standard deviation over three runs with different random seeds. The dimensions of the feature (penultimate layer output) space for all networks are set to 2048.}
    \label{tab:classifier}
\end{table}

\subsection{Evaluation on Far-OoD Benchmarks}

This subsection provides a detailed experimental evaluation of various OoD detection methods on two different Far-OoD benchmark datasets (Far-OoD (particles \& bubbles) and Far-OoD (General)). Far-OoD samples are crucial for evaluating the robustness of OoD detectors, as they represent data points that are semantically distinct from In-Distribution (ID) marine plankton samples. These samples include images that are highly unlikely to appear in real marine environments, such as general natural images unrelated to marine life, as well as objects that may exist in water but are far removed from our primary target, such as abiotic particles and bubbles. Effectively distinguishing such samples is critical in practical marine science applications, as it helps prevent false positives and ensures focus remains on relevant biological entities.

\noindent
\textbf{Experimental Details.}  We trained our networks using the ID data detailed in \cref{sec3: dataset}. To mitigate the effects of random variation, we conducted three separate training runs for each network architecture with different random seeds. Following the OpenOOD Guidelines \cite{zhang2023openood}, we trained three checkpoints for each network and then tested the OoD methods on them. The final results presented in \cref{tab:far_ood_performance} are based on the best-performing network for each method, selected for its superior overall AUROC performance across both Far-OoD benchmarks. Specifically, for each method, we chose the network whose average AUROC on both benchmarks was highest. The table reports the mean FPR95, FPR99, and AUROC values for each method, with a full breakdown including variance available in the Appendix \ref{nerwork result}.

\begin{table*}[htbp!]
    \centering
    \small
    \begin{tabularx}{\textwidth}{@{}l *{10}{Y} c @{}} % 在末尾添加一个c列用于Network
    \toprule
    \multirow{2}{*}{\parbox[c]{1.5cm}{%
        \centering
        \vspace{3ex}%
        \textbf{Method}%
    }} & \multicolumn{5}{c}{\textbf{Far-OoD(Bubbles \& Particles)}} & \multicolumn{5}{c}{\textbf{Far-OoD(General)}} & \multirow{2}{*}{\parbox[c]{1.5cm}{%
        \centering
        \vspace{3ex}%
        \textbf{Network}%
    }} \\
    \cmidrule(lr){2-6} \cmidrule(lr){7-11} % 部分横线
    & \textbf{FPR95-ID}↓ & \textbf{FPR95-OoD}↓ & \textbf{FPR99-ID}↓ & \textbf{FPR99-OoD}↓ & \textbf{AUROC}↑ & \textbf{FPR95-ID}↓ & \textbf{FPR95-OoD}↓ & \textbf{FPR99-ID}↓ & \textbf{FPR99-OoD}↓ & \textbf{AUROC}↑ & \\
    \midrule % 中间细线
        \multicolumn{12}{@{}c@{}}{\textit{Distance-based Methods}} \\ % 跨越11列，居中，斜体
        \addlinespace[0.5ex] % 增加一点垂直间距
        
        Mahalanobis & 21.44 & 11.90 & 61.01 & 22.96 & 96.67 & \textbf{0} & \textbf{0.03} & \textbf{0} & \textbf{0.04} & \textbf{99.98} & DenseNet-169 \\ 
        RMDS & 35.93 & 16.48 & 90.20 & 43.55 & 94.06 & 7.57 & 5.44 & 34.76 & 8.29 & 98.61 & DenseNet-201 \\
        KNN & 28.38 & 18.53 & 61.24 & 40.24 & 95.17 & 10.08 & 8.93 & 28.91 & 20.35 & 98.13 & ResNet-152 \\ 
        fDBD & 29.25 & 18.81 & 71.31 & 37.19 & 95.05 & 16.43 & 11.92 & 56.69 & 26.71 & 96.74 & DenseNet-201 \\ 
        
        \midrule % 中间细线
        \multicolumn{12}{@{}c@{}}{\textit{Classification-based Methods}} \\ % 跨越11列，居中，斜体
        \addlinespace[0.5ex] % 增加一点垂直间距
        ViM & \textbf{13.82} & \textbf{10.27} & \textbf{45.59} & \textbf{21.08} & \textbf{97.57} & 0.01 & 0.05 & 0.14 & 0.16 & 99.97 & DenseNet-201 \\ 
        Residual & 27.66 & 16.28 & 66.49 & 27.87 & 95.65 & \textbf{0} & 0.04 & 0.03 & 0.08 & 99.97 & DenseNet-169 \\
        ODIN$^{*}$ & 35.48 & 33.75 & 67.43 & 71.63 & 92.72 & 15.53 & 13.44 & 35.53 & 40.99 & 96.78 & SE-ResNeXt-50 \\  
        OpenMax & 74.93 & 24.07 & 95.99 & 48.37 & 90.45 & 30.42 & 20.34 & 67.87 & 49.95 & 94.62 & ResNet-152 \\ 
        Relation & 33.71 & 25.77 & 67.99 & 52.87 & 93.82 & 27.08 & 14.49 & 72.47 & 30.26 & 95.43 & DenseNet-201 \\
        TempScale & 39.90 & 31.04 & 68.63 & 70.99 & 92.19 & 51.98 & 35.46 & 82.56 & 69.11 & 89.77 & SE-ResNeXt-50 \\ 
        GEN & 37.19 & 32.20 & 67.05 & 72.50 & 92.41 & 48.29 & 37.56 & 84.11 & 71.34 & 89.77 & SE-ResNeXt-50 \\ 
        MSP & 37.32 & 22.16 & 71.26 & 61.67 & 93.54 & 47.38 & 60.33 & 82.25 & 84.20 & 87.58 & DenseNet-201 \\  
        MCDropout & 39.43 & 28.45 & 75.70 & 70.63 & 92.67 & 50.03 & 63.23 & 86.45 & 86.43 & 86.71 & DenseNet-201 \\
        MLS & 56.81 & 42.44 & 86.91 & 64.24 & 87.72 & 35.54 & 18.09 & 81.10 & 30.21 & 94.19 & ViT \\ 
        KL Matching & 36.80 & 66.07 & 72.12 & 91.81 & 89.94 & 41.88 & 60.20 & 73.63 & 80.89 & 87.57 & DenseNet-201 \\ 
        ReAct & 42.99 & 30.05 & 68.54 & 50.47 & 92.55 & 65.53 & 51.74 & 88.30 & 67.46 & 83.77 & DenseNet-201 \\  
        ASH & 40.61 & 36.37 & 77.14 & 60.53 & 91.89 & 73.21 & 74.00 & 94.72 & 85.51 & 74.20 & DenseNet-201 \\ 
        SHE & 79.53 & 72.57 & 93.28 & 83.48 & 72.04 & 49.6 & 51.64 & 75.52 & 64.27 & 85.21 & ViT \\ 
        RankFeat$^{\ddag}$ & 92.81 & 90.87 & 97.97 & 97.61 & 52.43 & 69.69 & 79.43 & 83.01 & 93.09 & 61.46 & ResNet-50 \\ 
        GradNorm & 66.89 & 71.40 & 88.15 & 90.22 & 79.57 & 32.88 & 29.79 & 68.84 & 55.30 & 92.79 & ViT \\
        
        \midrule % 中间细线
        \multicolumn{12}{@{}c@{}}{\textit{Density-based Methods}} \\ % 跨越11列，居中，斜体
        \addlinespace[0.5ex] % 增加一点垂直间距
        Energy & 57.44 & 42.73 & 87.94 & 64.10 & 87.53 & 36.48 & 18.22 & 83.46 & 30.12 & 94.05 & ViT \\
        DICE & 35.57 & 50.73 & 62.76 & 85.02 & 90.22 & 34.80 & 54.80 & 65.70 & 79.37 & 89.68 & SE-ResNeXt-50 \\
        \bottomrule % 底部粗线
    \end{tabularx}
    \caption{Comparision between the distance-based methods, classification-based method and density-based method on Far-OoD benchmark. All values are percentages. ↓ indicates smaller values are better and vice versa. For the Far-OoD(General) results, we take the average over the seven OoD test datasets it contains. The best metric is emphasized in bold. ODIN$^{*}$: Due to high computational cost and GPU memory limitations, we only tested this method on ResNet-18, ResNet-50, and SE-ResNeXt-50. RankFeat$^{\ddag}$: As this method requires intermediate layer features, we followed the OpenOOD implementation and tested it exclusively on the ResNet series and SE-ResNeXt networks.}
    \label{tab:far_ood_performance} % 添加一个标签，方便交叉引用
\end{table*}

\noindent
\textbf{Far-OoD Detection Performance.} In \cref{tab:far_ood_performance}, we compare the results of different methods on the Far-OoD benchmarks and highlight in \textbf{bold} the best-performing method. In total, distance-based methods significantly outperform classified-based and density-based methods on these benchmarks. Specifically, the Mahalanobis method achieves the best performance on the Far-OoD (General) benchmark, controlling both FPR95-ID and FPR99-ID to near zero. While Mahalanobis excels in this area, the ViM method demonstrates the most robust overall performance. ViM not only maintains a highly controlled FPR on the Far-OoD (General) benchmark but also effectively lowers the FPR on the more challenging Far-OoD (Bubbles \& Particles) benchmark. On this benchmark, ViM controls FPR95-ID and FPR99-ID to 13.82\% and 45.59\%, respectively, with an average AUROC of 97.57\%, which is a 4.03\% improvement in AUROC over the baseline MSP method. 

\noindent
\textbf{Comparison of General Baseline Methods.} Furthermore, we aimed to compare the performance of various baseline methods. As an example, we selected commonly used benchmark methods in Out-of-Distribution (OoD) detection: MSP, KNN, and Mahalanobis, each tested as a post hoc classifier. Our observations highlight the following:
\begin{itemize}
    \item \textbf{MSP vs. Mahalanobis.} Due to the potential for overconfident predictions in MSP \cite{nguyen2015deep}, its performance was not expected to be favorable. The results presented in  \cref{tab:far_ood_performance} corroborate this hypothesis. Compared to Mahalanobis, which demonstrated the best performance among the three methods, MSP exhibits increased values across FPR95-ID, FPR95-OoD, FPR99-ID, and FPR99-OoD for Far-OoD results, particularly for Far-OoD (General). This suggests that MSP struggles with samples that are entirely unrelated to the In-Distribution (ID) data and are significantly distant in the feature space.
    \item \textbf{Effectiveness of Feature Space for Separating ID and Far-OoD.} Distance-based methods (KNN and Mahalanobis) can directly leverage distance information within the feature space to assess the anomaly degree of samples. For Far-OoD samples, these methods effectively capture the absolute distance between the samples and the core ID distribution, thereby achieving robust discrimination. This aligns with their superior performance observed in both Far-OoD benchmarks.
\end{itemize}

\subsection{Evaluation on Near-OoD Benchmarks}

We further evaluated the performance of OoD detection tasks based on Near-OoD data. Compared to Far-OoD benchmarks, Near-OoD data is semantically closer to ID data and has fewer samples, making it more challenging as it requires higher model discrimination capabilities. We assessed the existing methods to identify those that can balance the performance of both Near-OoD and Far-OoD detection, thereby demonstrating greater robustness.

\begin{table*}[!htbp]
    \centering
    \small
    
    \begin{tabularx}{\textwidth}{l >{\Centering\arraybackslash}X >{\Centering\arraybackslash}p{2.5cm} >{\Centering\arraybackslash}X >{\Centering\arraybackslash}p{2.5cm} >{\Centering\arraybackslash}X >{\Centering\arraybackslash}X}
    \toprule % 顶部粗线
        \textbf{Method} & \textbf{FPR95-ID}↓ & \textbf{FPR95-OoD}↓ & \textbf{FPR99-ID}↓ & \textbf{FPR99-OoD}↓ & \textbf{AUROC}↑ & \textbf{Network}\\ 
        
        \midrule % 中间细线
        \multicolumn{7}{@{}c@{}}{\textit{Distance-based Methods}} \\ % 跨越11列，居中，斜体
        \addlinespace[0.5ex] % 增加一点垂直间距

        Mahalanobis & 44.58 & 21.09 & 82.60 & 34.60 & 93.40 & DenseNet-169  \\ 
        RMDS & 31.53 & 15.70 & 88.43 & 45.21 & 94.46 & DenseNet-121 \\ 
        KNN & 32.87 & 18.83 & 73.19 & 34.24 & 94.85 & ResNet-50 \\ 
        fDBD & 29.95 & 18.18 & 67.25 & 32.54 & 95.36 & DenseNet-169 \\ 
        
        \midrule % 中间细线
        \multicolumn{7}{@{}c@{}}{\textit{Classification-based Methods}} \\ % 跨越11列，居中，斜体
        \addlinespace[0.5ex] % 增加一点垂直间距
        ViM & \textbf{23.08} & \textbf{14.14} & 64.25 & \textbf{26.46} & \textbf{96.26} & DenseNet-169 \\
        Residual & 56.93 & 30.05 & 85.08 & 42.79 & 90.49 & DenseNet-169 \\
        ODIN$^{*}$ & 32.26 & 21.50 & 74.77 & 53.32 & 94.19 & ResNet-18 \\ 
        OpenMax & 89.04 & 17.32 & 99.5 & 34.39 & 90.35 & DenseNet-121 \\ 
        Relation & 34.24 & 23.61 & 67.89 & 36.14 & 94.15 & DenseNet-201 \\ 
        TempScale & 31.79 & 18.71 & 67.10 & 50.91 & 94.77 & DenseNet-121 \\ 
        GEN & 25.44 & 18.11 & 60.78 & 48.69 & 95.33 & DenseNet-121 \\ 
        MSP & 35.29 & 18.85 & 70.51 & 44.59 & 94.41 & DenseNet-121 \\ 
        MCDropout & 35.14 & 24.30 & 71.42 & 61.42 & 93.66 & DenseNet-169 \\
        MLS & 23.89 & 21.55 & 59.85 & 73.06 & 94.67 & DenseNet-121 \\ 
        KL Matching & 32.31 & 39.27 & 71.18 & 88.75 & 91.97 & DenseNet-169 \\ 
        ReAct & 31.38 & 26.45 & 65.18 & 50.54 & 93.72 & ResNet-18 \\ 
        ASH & 38.23 & 36.06 & 67.45 & 61.35 & 91.86 & DenseNet-121 \\ 
        SHE & 80.57 & 66.99 & 93.47 & 76.30 & 73.06 & ViT \\ 
        RankFeat$^{\ddag}$ & 89.07 & 88.13 & 97.14 & 97.01 & 62.27 & ResNet-18 \\ 
        GradNorm & 67.72 & 63.24 & 90.33 & 85.43 & 81.05 & ViT \\ 

        \midrule % 中间细线
        \multicolumn{7}{@{}c@{}}{\textit{Density-based Methods}} \\ % 跨越11列，居中，斜体
        \addlinespace[0.5ex] % 增加一点垂直间距
        Energy & 23.63 & 21.46 & \textbf{57.49} & 73.07 & 94.73 & DenseNet-121 \\ 
        DICE & 26.89 & 19.02 & 58.48 & 54.73 & 95.09 & ResNet-18 \\ 
        \bottomrule % 底部粗线
    \end{tabularx}
    \caption{Comparision between the distance-based methods, classification-based method and density-based method on Near-OoD benchmark. All values are percentages. ↓ indicates smaller values are better and vice versa. The best metric is emphasized in bold. }
    \label{tab:near-OoD results}
\end{table*}

\noindent
\textbf{Near-OoD Detection Performance.} In the Near-OoD benchmark evaluation, most detection methods showed improved performance, with a few exceptions among distance-based approaches. Notably, density-based methods like Energy and DICE proved highly effective at distinguishing these semantically similar anomalies, significantly reducing both FPR95 and FPR99 while substantially increasing AUROC. The ViM method maintained its superior overall performance, achieving an impressive AUROC of 96.26\%. This is attributed to ViM’s ability to leverage both discriminative information from the feature space and density-based insights from energy scores, allowing it to capture subtle distributional differences with exceptional precision.

\noindent
\textbf{Analysis of Method Specificity and Robustness.} Our analysis of the results across Far-OoD and Near-OoD benchmarks reveals that different detection methods exhibit significant specialization. Some methods, such as ViM and KNN, demonstrate strong generalization capabilities without requiring additional training, consistently maintaining high AUROC and low FPR values across both scenarios. This highlights their robustness and versatility. In contrast, other methods show a clear preference for specific OoD types. For instance, Residual excels at Far-OoD tasks but shows limited discriminative power for semantically closer Near-OoD samples. Conversely, density-based methods like Energy, DICE, and ReAct show superior performance in Near-OoD detection but may not be as effective for Far-OoD tasks. This underscores the critical importance of selecting a detection strategy tailored to the specific characteristics of the OoD data in a given application, especially in fields like plankton detection where precise identification of both novel and rare categories is essential \cite{tajwar2021no}.
 
\noindent
\textbf{Performance Insight for Distance-Based Methods.}  \Cref{tab:far_ood_performance} and \Cref{tab:near-OoD results} reveal that for distance-based methods, FPR-ID is typically greater than FPR-OoD. This phenomenon may stem from ID data being highly centralized in their feature space. By compressing known category samples into tight core regions, these models effectively identify and exclude true OoD samples. This holds even for semantically similar Near-OoD instances, significantly reducing false positives for OoD. However, this strategy can lead to overly strict judgment of ID data itself. Consequently, marginal or less typical ID samples may be erroneously classified as OoD, which in turn elevates the FPR-ID.

% \noindent
% \textbf{Impact of Network Architecture.}
% We conducted an ablation study to investigate the influence of network architecture on method performance (Appendix \ref{ablation study}). We observed that some methods, such as GradNorm, ReAct, ASH, and SHE, exhibit strong dependence on the underlying network, while others, including KNN, fDBD, Relation, and ViM, are less sensitive. This highlights the importance of considering the chosen network architecture when evaluating OoD detection results.
% 网络架构对方法的影响
% 我们做了一项消融实验，探究网络架构对方法表现的影响。我们发现，有的方法效果严重依赖于使用的网络，如GradNorm、ReAct、ASH、SHE等；相反，有的方法对使用网络并不那么敏感，如KNN、fDBD、Relation、ViM等。这提示我们需要考虑到采用的网络架构对于结果的影响。
\section{Discussion and Conclusions}
\label{sec5:conclution}

% 基于我们的研究结果，我们发现现有 OoD 检测方法在浮游生物检测这一特定应用场景中展现出显著潜力。然而，将这些方法从通用数据集扩展到实际海洋生态监测任务时，我们观察到仍存在一些关键挑战。首先，浮游生物物种之间常常存在高度的形态相似性，导致其语义差异不足，这使得在更细致的粒度上进行特征检测和区分变得尤为关键。其次，同一物种内部也可能因生命周期、环境影响等因素呈现显著的形态差异，而不同地理位置或时间采集的样本，即使属于同一类别，其视觉特征也可能大相径庭，这些都增加了 OoD 检测的复杂性。此外，由于不同采集系统获取的图像特征差异显著，且可能存在噪声、模糊等数据质量不均的问题，将直接影响检测模型的性能。同时，不同浮游生物物种在自然环境中的出现频率差异巨大，导致数据集中类别分布严重不平衡，这极大地挑战了对稀有物种的准确识别能力。

Based on our research findings, we observe a significant potential for existing OoD detection methods in the specific application scenario of plankton detection. However, extending these methods from general datasets to real-world marine ecological monitoring tasks presents several key challenges. Firstly, plankton species often exhibit high morphological similarity, leading to insufficient semantic clarity among different categories, which makes fine-grained feature detection and differentiation particularly crucial. Secondly, significant morphological variations can exist within the same species due to life cycles or environmental influences, and samples collected from different geographical locations or times, even if belonging to the same category, may show substantial visual disparities. These factors collectively increase the complexity of OoD detection \cite{ciranniComputerVisionDeep2024, bachimanchi2024deep, eerola2024survey}. Furthermore, varying image features acquired from different collection systems, coupled with potential issues like noise and blur, result in uneven data quality that directly impacts detection model performance. Simultaneously, the vast differences in natural occurrence frequencies among different plankton species lead to severely imbalanced class distributions in datasets, posing a significant challenge to the accurate identification of rare species \cite{ciranniComputerVisionDeep2024, eerola2024survey}.

% 鉴于上述挑战，为了提高开放场景下浮游生物检测模型的可靠性，我们认为以下方向的进一步探索将显著提升 OoD 检测模型的性能：首先，本研究验证了事后（post-hoc）方法的有效性，这些方法无需额外的训练过程。这对于应对实际海洋监测中数据质量不均和类别不平衡的问题，避免大规模数据收集和模型重训练的昂贵成本，具有显著的应用潜力。因此，此类方法在未来浮游生物图像分析中值得深入探索。其次，在实际浮游生物检测任务中，针对物种间的高度形态相似性和近距离 OoD 样本的区分难题，有时需要在微小尺度上区分分布内（ID）和 OoD 实例，例如区分形态相似的浮游生物物种或将其与非生物颗粒区分开来，这要求从细粒度分类的角度进一步提取判别性特征以支持 OoD 检测。最后，考虑到浮游生物图像中存在的形态变异和可能的多物种混合现象，开发适用于多标签分类的 OoD 检测方法，将有助于处理大规模、多样的浮游生物群落检测任务，从而全面提升模型的鲁棒性。
 Given these challenges, to enhance the reliability of plankton detection models in open-set scenarios, we believe that further exploration in the following directions will significantly improve OoD detection model performance: Firstly, this study validates the effectiveness of post hoc methods, which do not necessitate additional training processes. This is particularly beneficial for addressing issues of uneven data quality and class imbalance in real-world marine monitoring, avoiding the costly burden of large-scale data collection and model retraining. Thus, such methods warrant deeper investigation for future plankton image analysis. Secondly, in practical plankton detection tasks, to address the high morphological similarity between species and the difficulty in distinguishing Near-OoD samples, it is sometimes necessary to differentiate ID and OoD instances at a minute scale, for example, distinguishing between morphologically similar plankton species or separating them from non-biological particles. This requires further extraction of discriminative features from a fine-grained classification perspective to support OoD detection. Lastly, considering the morphological variations and potential mixed phenomena present in plankton imagery, developing OoD detection methods suitable for multi-label classification would be beneficial for handling large-scale, diverse plankton community detection tasks, consequently enhancing overall model robustness.

% 综上所述，为了提高浮游生物检测模型的可靠性和安全性，我们对一系列具有高度代表性的 OoD 检测方法进行了全面评估。为更深入地比较不同方法在形态语义相似性和环境变化下的性能，我们在 DYB-PlanktonNet 数据集上精心构建了兼顾 Near-OoD 和 Far-OoD 的基准测试，并使用 AUROC、FPR95 和 FPR99 等指标进行了定量评估。通过大量的实验，我们发现 ViM 方法在无需额外训练的前提下，在所有 OoD 基准测试上均展现出卓越的综合性能，尤其在平衡 Far-OoD 和 Near-OoD 检测方面表现突出。这有力地表明，额外训练并非浮游生物图像检测应用中提升 OoD 性能的唯一途径。我们的研究结果不仅证明了现有 OoD 检测方法能够为大规模浮游生物检测部署提供可靠性和安全性，即使面对多样的形态覆盖范围和复杂的环境条件，同时也为未来探索更适合大规模浮游生物检测应用的 OoD 检测方法提供了宝贵的见解和指引。
In summary, to improve the reliability and robustness of plankton detection models, we conducted a comprehensive evaluation of a set of highly representative OoD detection methods. To further compare the performance of various methods under morphological semantic similarity and environmental variations, we meticulously constructed a series of benchmarks on the DYB-PlanktonNet dataset, encompassing both Near-OoD and Far-OoD, and quantitatively evaluated them using AUROC, FPR95, and FPR99 metrics. Through extensive experimentation, we found that the ViM method demonstrated excellent comprehensive performance across all OoD benchmarks, notably excelling in balancing both Far-OoD and Near-OoD detection. Our findings not only demonstrate that existing OoD detection methods can provide reliability and safety for large-scale plankton detection deployments, even when faced with diverse morphological coverages and complex environmental conditions, but also offer valuable insights and guidance for future exploration of OoD detection methods better suited for large-scale plankton detection applications.

\section*{Acknowledgements}

This work was supported in part by the National Nature Science Foundation of China (No.12201048), National Natural Science Foundation of China (No.42476218). The authors thank support from the Interdisciplinary Intelligence Super Computer Center of Beijing Normal University at Zhuhai.

{
    \small
    \bibliographystyle{ieeenat_fullname}
    \bibliography{main}
}

% WARNING: do not forget to delete the supplementary pages from your submission 
\clearpage
\setcounter{page}{1}
\maketitlesupplementary

\setcounter{section}{0} 

\section{Dataset Detailed Categories}
\label{sec:dataset}
\setlength{\belowcaptionskip}{-0.2cm}

% 本节提供了我们为评估异常数据（Out-of-Distribution, OoD）检测方法而构建的浮游生物数据集的详细分类信息。为了模拟真实世界中海洋生态监测面临的各种分布漂移情景，我们将DYB-PlanktonNet数据集中的92个原始类别细致地划分为三个子集：In-Distribution (ID)、Near-OoD和Far-OoD。这种分层分类方法旨在对不同语义和形态相似度的异常数据进行精确评估，从而更全面地反映模型在实际部署中的性能表现。Tab 4,5,6 详细列出了每个子集中的所有类别，以及它们的具体含义和在我们的基准测试中的作用。

This section provides detailed classification information for the plankton dataset we constructed to evaluate Out-of-Distribution (OoD) detection methods. To simulate various distribution shift scenarios encountered in real-world marine ecological monitoring, we meticulously divided the ninety-two original classes from the DYB-PlanktonNet dataset into three subsets: In-Distribution (ID), Near-OoD, and Far-OoD. This hierarchical classification approach is designed to accurately evaluate anomalous data with varying semantic and morphological similarities, thus more comprehensively reflecting the model's performance in practical deployment. \Cref{tab:ID,tab:Near-OoD,tab:Far-OoD} provide a detailed list of all categories in each subset, along with their specific meanings and roles in our benchmark.

\begin{table*}[!htbp]
    \centering
    \footnotesize
    \begin{tabular}{ccccc}
    \hline
        \textbf{ID-class} & \textbf{Specimen type} & \textbf{Phylum} & \textbf{Class} & \textbf{Order} \\ \hline
        Polychaeta\_most with eggs & Plankton & Annelida & Polychaeta & / \\ 
        Polychaeta\_Type A & Plankton & Annelida & Polychaeta & / \\ 
        Polychaeta\_Type B & Plankton & Annelida & Polychaeta & / \\ 
        Polychaeta\_Type C & Plankton & Annelida & Polychaeta & / \\ 
        Polychaeta\_Type D & Plankton & Annelida & Polychaeta & / \\ 
        Polychaeta\_Type E & Plankton & Annelida & Polychaeta & / \\ 
        Polychaeta\_Type F & Plankton & Annelida & Polychaeta & / \\ 
        Penilia avirostris & Plankton & Arthropoda & Branchiopoda & Ctenopoda  \\ 
        Evadne tergestina & Plankton & Arthropoda & Branchiopoda  & Onychopoda \\ 
        Acartia sp.A & Plankton & Arthropoda & Hexanauplia & Calanoida  \\ 
        Acartia sp.B & Plankton & Arthropoda & Hexanauplia & Calanoida  \\ 
        Acartia sp.C & Plankton & Arthropoda & Hexanauplia & Calanoida  \\ 
        Calanopia sp. & Plankton & Arthropoda & Hexanauplia & Calanoida  \\ 
        Labidocera sp. & Plankton & Arthropoda & Hexanauplia & Calanoida  \\ 
        Tortanus gracilis & Plankton & Arthropoda & Hexanauplia & Calanoida  \\ 
        Calanoid with egg & Plankton & Arthropoda & Hexanauplia & Calanoida  \\ 
        Calanoid\_Type A & Plankton & Arthropoda & Hexanauplia & Calanoida  \\ 
        Calanoid\_Type B & Plankton & Arthropoda & Hexanauplia & Calanoida  \\ 
        Oithona sp.B with egg & Plankton & Arthropoda & Hexanauplia & Cyclopoida \\ 
        Cyclopoid\_Type A\_with egg & Plankton & Arthropoda & Hexanauplia & Cyclopoida \\ 
        Harpacticoid\_mating & Plankton & Arthropoda & Hexanauplia & Harpacticoida \\ 
        Microsetella sp. & Plankton & Arthropoda & Hexanauplia & Harpacticoida  \\ 
        Caligus sp. & Plankton & Arthropoda & Hexanauplia & Siphonostomatoida \\ 
        Copepod\_Type A & Plankton & Arthropoda & Hexanauplia & / \\ 
        Caprella sp. & Plankton & Arthropoda & Malacostraca & Amphipoda \\ 
        Amphipoda\_Type A & Plankton & Arthropoda & Malacostraca & Amphipoda \\ 
        Amphipoda\_Type B & Plankton & Arthropoda & Malacostraca & Amphipoda \\ 
        Amphipoda\_Type C & Plankton & Arthropoda & Malacostraca & Amphipoda \\ 
        Gammarids\_Type A & Plankton & Arthropoda & Malacostraca & Amphipoda \\ 
        Gammarids\_Type B & Plankton & Arthropoda & Malacostraca & Amphipoda \\ 
        Gammarids\_Type C & Plankton & Arthropoda & Malacostraca & Amphipoda \\ 
        Cymodoce sp.  & Plankton & Arthropoda & Malacostraca & Isopoda \\ 
        Lucifer sp. & Plankton & Arthropoda & Malacostraca & Decapoda  \\ 
        Macrura larvae & Plankton & Arthropoda & Malacostraca & Decapoda  \\ 
        Megalopa larva\_Phase 1\_Type B & Plankton & Arthropoda & Malacostraca & Decapoda  \\ 
        Megalopa larva\_Phase 1\_Type C & Plankton & Arthropoda & Malacostraca & Decapoda  \\ 
        Megalopa larva\_Phase 1\_Type D & Plankton & Arthropoda & Malacostraca & Decapoda  \\ 
        Megalopa larva\_Phase 2 & Plankton & Arthropoda & Malacostraca & Decapoda  \\ 
        Porcrellanidae larva & Plankton & Arthropoda & Malacostraca & Decapoda  \\ 
        Shrimp-like larva\_Type A & Plankton & Arthropoda & Malacostraca & Decapoda  \\ 
        Shrimp-like larva\_Type B & Plankton & Arthropoda & Malacostraca & Decapoda  \\ 
        Shrimp-like\_Type A & Plankton & Arthropoda & Malacostraca & Decapoda  \\ 
        Shrimp-like\_Type B & Plankton & Arthropoda & Malacostraca & Decapoda  \\ 
        Shrimp-like\_Type D & Plankton & Arthropoda & Malacostraca & Decapoda  \\ 
        Shrimp-like\_Type F & Plankton & Arthropoda & Malacostraca & Decapoda  \\ 
        Cumacea\_Type A & Plankton & Arthropoda & / & / \\ 
        Cumacea\_Type B & Plankton & Arthropoda & / & / \\ 
        Chaetognatha & Plankton & Chaetognatha & / & / \\ 
        Oikopleura sp. parts & Plankton & Chordata & Appendicularia & Copelata \\ 
        Tunicata\_Type A & Plankton & Chordata & / & / \\ 
        Jellyfish & Plankton & Cnidaria & / & / \\ 
        Creseis acicula & Plankton & Mollusca & Gastropoda & Pteropoda \\ 
        Noctiluca scintillans & Plankton & Myzozoa & Dinophyceae & Noctilucales \\ 
        Phaeocystis globosa & Plankton & Haptophyta & / & / \\ \hline
    \end{tabular}
    \caption{In-Distribution (ID) Class}
    \label{tab:ID}
\end{table*}

\begin{table*}[!htbp]
    \centering
    \begin{tabular}{ccccc}
    \hline
        \textbf{Near-OoD-class} & \textbf{Specimen type} & \textbf{Phylum} & \textbf{Class} & \textbf{Order} \\ \hline
        Polychaeta larva & Plankton & Annelida & Polychaeta & / \\ 
        Calanoid Nauplii & Plankton & Arthropoda & Hexanauplia & Calanoida  \\ 
        Calanoid\_Type C & Plankton & Arthropoda & Hexanauplia & Calanoida  \\ 
        Calanoid\_Type D & Plankton & Arthropoda & Hexanauplia & Calanoida  \\ 
        Oithona sp.A with egg & Plankton & Arthropoda & Hexanauplia & Cyclopoida \\ 
        Cyclopoid\_Type A & Plankton & Arthropoda & Hexanauplia & Cyclopoida \\ 
        Harpacticoid & Plankton & Arthropoda & Hexanauplia & Harpacticoida \\ 
        Monstrilla sp.A & Plankton & Arthropoda & Hexanauplia & Monstrilloida \\ 
        Monstrilla sp.B & Plankton & Arthropoda & Hexanauplia & Monstrilloida \\ 
        Megalopa larva\_Phase 1\_Type A & Plankton & Arthropoda & Malacostraca & Decapoda  \\ 
        Shrimp-like\_Type C & Plankton & Arthropoda & Malacostraca & Decapoda  \\ 
        Shrimp-like\_Type E & Plankton & Arthropoda & Malacostraca & Decapoda  \\ 
        Ostracoda & Plankton & Arthropoda & Ostracoda & / \\ 
        Oikopleura sp. & Plankton & Chordata & Appendicularia & Copelata \\ 
        Actiniaria larva & Plankton & Cnidaria & Anthozoa & / \\ 
        Hydroid & Plankton & Cnidaria & / & / \\ 
        Jelly-like & Plankton & Cnidaria & / & / \\ 
        Bryozoan larva & Plankton & Ectoprocta/bryozoan & / & / \\ 
        Gelatinous Zooplankton & Plankton & / & / & / \\ 
        Unknown\_Type A & Plankton & / & / & / \\ 
        Unknown\_Type B & Plankton & / & / & / \\ 
        Unknown\_Type C & Plankton & / & / & / \\ 
        Unknown\_Type D & Plankton & / & / & / \\ 
        Balanomorpha exuviate & Carcass & Arthropoda & Hexanauplia & Sessilia  \\ 
        Monstrilloid & Plankton & Arthropoda & Hexanauplia & Monstrilloida \\ 
        Fish Larvae & Chordata & Vertebrata & Actinopterygii & / \\ \hline
    \end{tabular}
    \caption{Near-OoD Class}
    \label{tab:Near-OoD}
\end{table*}

\begin{table*}[!htbp]
    \centering
    \begin{tabular}{cccc}
    \hline
        \textbf{Far-OoD-class} & \textbf{Specimen type} & \textbf{Phylum} & \textbf{Class} \\ \hline
        Crustacean limb\_Type A & Carcass & Arthropoda & / \\ 
        Crustacean limb\_Type B & Carcass & Arthropoda & / \\ 
        Fish egg & Chordata & Vertebrata & Actinopterygii \\ 
        Particle\_filamentous\_Type A & Unknown & / & / \\ 
        Particle\_filamentous\_Type B & Non-Living & / & / \\ 
        Particle\_bluish & Non-Living & / & / \\ 
        Particle\_molts & Non-Living & / & / \\ 
        Particle\_translucent flocs & Non-Living & / & / \\ 
        Particle\_yellowish flocs & Non-Living & / & / \\ 
        Particle\_yellowish rods & Non-Living & / & / \\ 
        Bubbles & Non-Living & / & / \\ 
        Fish tail & Non-Living & / & / \\ \hline
    \end{tabular}
    \caption{Far-OoD (Bubbles \& Particles) Class}
    \label{tab:Far-OoD}
\end{table*}

\section{Common OoD post hoc methods}
\label{oodmethods_intro}
%表x介绍了我们实验的OoD检测方法的基本原理。

\Cref{tab:ood-methods} outlines the basic principles of the OoD detection methods employed in our study.

\begin{table*}[hb]
    \centering
    \fontsize{8.5}{11.5}\selectfont
    \begin{tabularx}{\textwidth}{@{} l Y Y @{}}
    \toprule
    \textbf{Method} & \textbf{Score Function} & \textbf{Note} \\
    \midrule

    \multicolumn{3}{@{}c@{}}{\textit{Distance-based Methods}} \\
    \addlinespace[0.8ex]

    Mahalanobis
      & $\displaystyle -(\mathbf{z}-\mu_c)^\top \Sigma^{-1}(\mathbf{z}-\mu_c)$
      & Negative Mahalanobis distance to class-$c$ prototype ($\mu_c,\Sigma$ from training) \\
    \addlinespace[0.8ex]

    RMDS
      & $ -\min_{c}\bigl[(\mathbf{z}-\mu_c)^\top\Sigma_c^{-1}(\mathbf{z}-\mu_c) - (\mathbf{z}-\mu_0)^\top\Sigma_0^{-1}(\mathbf{z}-\mu_0)\bigr]$
      & Uses $\mu_0, \Sigma_0$ of entire training data as background \\
    \addlinespace[0.8ex]

    KNN
      & $\displaystyle -\lVert\mathbf{z} - \mathbf{z}_{(k)}\rVert_2$
      & $\mathbf{z}_{(k)}$ is the $k$th nearest inlier feature (features are normalized) \\
    \addlinespace[0.8ex]

    fDBD
      & $\displaystyle -\frac{1}{\lvert C\rvert -1}\sum_{c\neq y}\frac{\tilde D_f(\mathbf{z},c)}{\lVert\mathbf{z}-\mu_{\mathrm{train}}\rVert_2}$
      & $\tilde D_f(\mathbf{z},c)=\frac{\lvert(\mathbf{w}_{y}-\mathbf{w}_c)^\top\mathbf{z}+(b_{y}-b_c)\rvert}{\lVert\mathbf{w}_{y}-\mathbf{w}_c\rVert_2}$, $y$ is predicted class, $\mathbf{W}=[\mathbf{w}_1,\cdots,\mathbf{w}_C]$ classifier weights, $\mu_{\mathrm{train}}$ training-feature mean \\

    \midrule

    \multicolumn{3}{@{}c@{}}{\textit{Classification-based Methods}} \\
    \addlinespace[0.8ex]

    ViM
      & $\displaystyle   -\alpha\lVert\mathbf{z}^{P^\perp}\rVert_2 + \log\sum_c e^{f_c(\mathbf{z})}$
      & Combines residual with LSE of logits $f_c(\mathbf{z})$ \\
    \addlinespace[0.8ex]

    Residual
      & $\displaystyle  -\lVert\mathbf{z}^{P^\perp}\rVert_2$
      & $\mathbf{z}^{P^\perp}$ is projection residual outside principal subspace \\
    \addlinespace[0.8ex]

    ODIN
      & $\displaystyle \max_c \sigma_{\mathrm{SM}}(f(\tilde{\mathbf{x}})/T)^{(c)}$
      & Perturb input $\tilde{\mathbf{x}} = \mathbf{x} + \varepsilon\,\mathrm{sign}\bigl(\nabla_{\mathbf{x}}\log p_{\max}(\mathbf{x})\bigr)$, then apply temp $T$-scaled softmax (operates in input space) \\
    \addlinespace[2ex]

    OpenMax
      & $\displaystyle \max_c \hat P(y=c \mid \mathbf{x})$
      & $\hat P(y=c \mid \mathbf{x})$ is recalibrated probability; accept if $\arg\max_j\hat P(y{=}j\mid\mathbf{x})\neq\text{unknown}$ (operates in input space) \\
    \addlinespace[2ex]

    TempScale
      & $\displaystyle   \max_c \sigma_{\mathrm{SM}}(f(\mathbf{z})/T)^{(c)}$
      & $\sigma_{\mathrm{SM}}$ is softmax with temperature $T$ \\
    \addlinespace[0.8ex]

    GEN
      & $\displaystyle G_\gamma(\mathbf{p}) = -\sum_{m=1}^C p_{i_m}^\gamma(1-p_{i_m})^\gamma$
      & $p_{i_1}\ge\cdots\ge p_{i_C}$ are sorted softmax probabilities, $\gamma\in(0,1)$ \\
    \addlinespace[0.8ex]

    MSP
      & $\displaystyle   \max_{c} p_c(\mathbf{z})$
      & Maximum softmax probability \\
    \addlinespace[0.8ex]

    MCDropout
      & $\displaystyle   -H\bigl(\tfrac{1}{T}\sum_{t=1}^T\hat{\mathbf{y}}^{(t)}(\mathbf{x})\bigr)$
      & $H(\cdot)$ is entropy of predictive mean over $T$ dropout samples (operates in input space) \\
    \addlinespace[0.8ex]

    MLS
      & $\displaystyle S_1(\mathbf{z})=\max_c f_c(\mathbf{z})$
      & MaxLogit \\
    \addlinespace[0.8ex]

    KL Matching
      & $\displaystyle   -\min_c D_{\mathrm{KL}}\bigl(\mathbf{p}(\mathbf{x})\parallel \mathbf{d}_c\bigr)$
      & $\mathbf{d}_c$ is class-prototype distribution (operates in input space) \\
    \addlinespace[0.8ex]

    ReAct
      & $\displaystyle \max_c \sigma_{\mathrm{SM}}(f(\min(\mathbf{z}, b))^{(c)}$
      & Clamp activations at threshold $b$ and apply MSP score \\
    \addlinespace[0.8ex]

    ASH
      & $\displaystyle   \log\sum_{c=1}^C \exp\bigl(f_c^{\mathrm{ASH}}(\mathbf{z})\bigr)$
      & $f^{\mathrm{ASH}}=\mathbf{W}^\top\mathbf{h}'(\mathbf{z})+\mathbf{b}$, $\mathbf{W}$ classifier weights, $\mathbf{h}'(\mathbf{z})$ is processed feature (pruning \& normalization) \\
    \addlinespace[0.8ex]

    SHE
      & $\displaystyle   \beta^{-1}\log\sum_{j=1}^M\exp\bigl(\beta\,\boldsymbol{\xi}^\top\mathbf{S}_j\bigr)$
      & $\beta$ is hyper-parameter, $\boldsymbol{\xi}^\mathrm{T} \mathbf{S}_j$ is inner product between test pattern and stored pattern \\
    \addlinespace[2ex]

    RankFeat
      & $\displaystyle \max_c f_c(\mathbf{z} - s_1\,\mathbf{u}_1\mathbf{v}_1^\top)$
      & Remove first principal component and apply MaxLogit \\
    \addlinespace[2ex]

    GradNorm
      & $\displaystyle   \lVert\mathbf{p}-\tfrac{1}{C}\mathbf{1}\rVert_1\cdot\lVert\mathbf{z}\rVert_1$
      & L1 distance of $\mathbf{p}$ to uniform distribution $(\times)$ feature norm \\
    \addlinespace[2ex]

    Relation
      & $\displaystyle   \sum_{i\in S} k(\mathbf{z},\mathbf{z}_i)$
      & $k(\cdot,\cdot)$ similarity kernel, $S$ support set of stored inlier features \\

    \midrule

    \multicolumn{3}{@{}c@{}}{\textit{Density-based Methods}} \\
    \addlinespace[0.8ex]

    Energy
      & $\displaystyle   T\log\sum_{c=1}^C\exp\bigl(f_c(\mathbf{z})/T\bigr)$
      & $f_c(\mathbf{z})$ is logit value, $T$ temperature \\
    \addlinespace[0.8ex]

    DICE
      & $\displaystyle   \log\sum_{c=1}^C\exp\bigl(((\mathbf{M}\odot\mathbf{W})^\top\mathbf{z})_c + b_c\bigr)$
      & $\mathbf{W}$ classifier weights, $\mathbf{M}$ mask matrix for sparsification \\
    \bottomrule
    \end{tabularx}
    \caption{Method Introduction}
    \label{tab:ood-methods}
\end{table*}

\section{Experiment Details}

% Dataset Preprocessing: 我们将 ID 数据集按照 8:1:1 的比例切分为训练集（train）、验证集（validation）和测试集（test）。骨干网络在训练集上进行训练，并在验证集上进行超参数调整。ID 类的分类准确率（ACC）则在测试集上进行计算。对于所有图像，我们均进行了归一化处理。在训练阶段，我们采用随机裁剪（random crop）和随机水平翻转（random horizontal flip）进行数据增强，以提高模型的泛化能力。验证和测试阶段，图像首先被调整大小，然后进行中心裁剪（center crop）。所有裁剪后的图片尺寸均为 224×224 像素，以作为网络输入。

\subsection{Dataset Preprocessing}  The ID dataset was split into training, validation, and testing subsets in a ratio of 8:1:1. All backbone networks were trained on the training split, while hyperparameter tuning was performed on the validation split. The classification accuracy (ACC) for ID classes was evaluated on the test split. All images underwent normalization as a preprocessing step. During training, we applied random cropping and random horizontal flipping for data augmentation to enhance model generalization. In the validation and testing phases, images were first resized and then subjected to center cropping. Consistent with the OpenOoD benchmark \cite{zhang2023openood}, our training protocol uses only standard data augmentation, without any advanced strategies. All cropped images were resized to a fixed resolution of 224×224 pixels before being fed into the network.

% 超参数搜索：考虑到 OoD 检测方法的性能往往对超参数选择非常敏感，为了实现公平且可复现的评估，我们严格遵循 OpenOoD 基准的设置。对于所有需要超参数调优的方法，我们都进行了详尽的超参数搜索，以报告其最佳性能。

\subsection{Hyperparameter Search} Given the high sensitivity of Out-of-Distribution (OoD) detection methods to hyperparameter choices, we adopted the OpenOoD-v1.5 Guidelines \cite{zhang2023openood} for a fair and reproducible evaluation. Specifically, we used a validation set to tune the hyperparameters for each method and backbone model. For all methods requiring tuning, we conducted an extensive hyperparameter search to determine their optimal settings. To account for randomness, this search was performed for each of the three separate training runs (with different random seeds). The specific hyperparameter values that yielded the best performance for each combination are detailed in \cref{tab:hyperparam}.

\begin{table*}[!htbp]
    \centering
    \footnotesize
    \begin{tabular}{cccccccccccc} 
    \hline
        % Network & seed & ASH & fDBD & GEN & ~ & KNN & ReAct & Relation & ViM & ODIN & ~ \\ 
        % ~ & ~ & percentile & distance\_as\_normalizer & gamma & M & K & percentile & pow & dim & temperature & noise \\
        \multicolumn{2}{c}{\textbf{Network}} & \multicolumn{8}{c}{\textbf{Hyperparameters}} \\
        \cmidrule(lr){3-12}
        \textbf{Backbone} & \textbf{Seed} & \textbf{ASH} & \textbf{fDBD} & \multicolumn{2}{c}{\textbf{GEN}} & \textbf{KNN} & \textbf{ReAct} & \textbf{Relation} & \textbf{ViM} & \multicolumn{2}{c}{\textbf{ODIN$^{*}$}} \\
        ~ & ~ & percentile & distance\_as\_normalizer & gamma & M & K & percentile & pow & dim & temperature & noise \\
        \hline
        \multirow{3}{*}{ResNet-18} & s0 & 95 & FALSE & 0.01 & 50 & 50 & 99 & 8 & 64 & 1 & 0.0014 \\ 
         & s1 & 95 & FALSE & 0.5 & 100 & 50 & 99 & 8 & 256 & 1 & 0.0014 \\ 
         & s2 & 95 & FALSE & 0.1 & 50 & 50 & 99 & 8 & 256 & 1 & 0.0014 \\ \hline
        \multirow{3}{*}{ResNet-50} & s0 & 95 & TRUE & 0.01 & 10 & 50 & 99 & 8 & 256 & 1 & 0.0014 \\ 
         & s1 & 95 & FALSE & 0.1 & 50 & 50 & 99 & 8 & 256 & 1 & 0.0014 \\ 
         & s2 & 95 & FALSE & 0.01 & 10 & 50 & 99 & 8 & 256 & 1 & 0.0014 \\ \hline
        \multirow{3}{*}{ResNet-101} & s0 & 95 & FALSE & 0.1 & 50 & 50 & 99 & 8 & 256 & ~ & ~ \\ 
         & s1 & 95 & FALSE & 0.5 & 50 & 50 & 99 & 8 & 256 & ~ & ~ \\ 
         & s2 & 95 & FALSE & 0.01 & 10 & 50 & 99 & 8 & 256 & ~ & ~ \\ \hline
        \multirow{3}{*}{ResNet-152} & s0 & 95 & TRUE & 0.01 & 10 & 50 & 99 & 8 & 256 & ~ & ~ \\ 
        ~ & s1 & 95 & FALSE & 0.5 & 50 & 50 & 99 & 8 & 256 & ~ & ~ \\ 
        ~ & s2 & 95 & FALSE & 0.1 & 50 & 50 & 99 & 8 & 256 & ~ & ~ \\ \hline
        \multirow{3}{*}{DenseNet-121} & s0 & 95 & FALSE & 0.01 & 10 & 50 & 99 & 8 & 128 & ~ & ~ \\ 
         & s1 & 95 & FALSE & 0.01 & 10 & 50 & 99 & 8 & 256 & ~ & ~ \\ 
         & s2 & 95 & FALSE & 0.1 & 50 & 50 & 99 & 8 & 256 & ~ & ~ \\ \hline
        \multirow{3}{*}{DenseNet-169} & s0 & 95 & FALSE & 0.01 & 50 & 50 & 99 & 8 & 256 & ~ & ~ \\ 
         & s1 & 95 & FALSE & 0.1 & 50 & 50 & 99 & 8 & 256 & ~ & ~ \\ 
         & s2 & 95 & FALSE & 0.01 & 10 & 50 & 99 & 8 & 256 & ~ & ~ \\ \hline
        \multirow{3}{*}{DenseNet-201} & s0 & 95 & FALSE & 0.01 & 10 & 50 & 99 & 8 & 256 & ~ & ~ \\ 
         & s1 & 95 & FALSE & 0.01 & 10 & 50 & 99 & 8 & 256 & ~ & ~ \\ 
         & s2 & 95 & FALSE & 0.01 & 10 & 50 & 99 & 8 & 256 & ~ & ~ \\ \hline
        \multirow{3}{*}{Se-ResNeXt-50} & s0 & 95 & FALSE & 0.01 & 10 & 50 & 99 & 8 & 256 & 1 & 0.0014 \\ 
         & s1 & 95 & FALSE & 0.01 & 10 & 50 & 99 & 8 & 256 & 1 & 0.0014 \\ 
         & s2 & 95 & FALSE & 0.01 & 10 & 50 & 99 & 8 & 256 & 1 & 0.0014 \\ \hline
        \multirow{3}{*}{ViT} & s0 & 95 & TRUE & 0.1 & 10 & 50 & 99 & 8 & 256 & ~ & ~ \\ 
         & s1 & 65 & TRUE & 0.1 & 50 & 50 & 99 & 8 & 256 & ~ & ~ \\ 
         & s2 & 80 & TRUE & 0.1 & 10 & 50 & 99 & 8 & 256 & ~ & ~ \\ \hline
    \end{tabular}
    \caption{Optimal Hyperparameters for OoD Detection Methods. This table lists the best-performing hyperparameter configurations found for each backbone network and OoD detection method after an hyperparameter search. \textbf{ODIN*} was only evaluated on the ResNet-18, ResNet-50, and Se-ResNeXt-50 backbones due to its significant computational cost.}
    \label{tab:hyperparam}
\end{table*}

\subsection{Ablation Study}
\label{ablation study}

% 为了讨论网络架构对OoD检测方法的影响，我们做了一项消融实验，仅替换了network backbone。每个网络我们使用不同的随机种子进行了3次重复实验，并汇报了在Near-OoD、Far-OoD(Bubbles & Perticles)和Far-OoD(General)上的AUROC值。对于一些需要最优超参数的方法，我们进行了超参数搜索。结果如图所示。

\begin{figure*}
    \centering
    \includegraphics[width=1\linewidth]{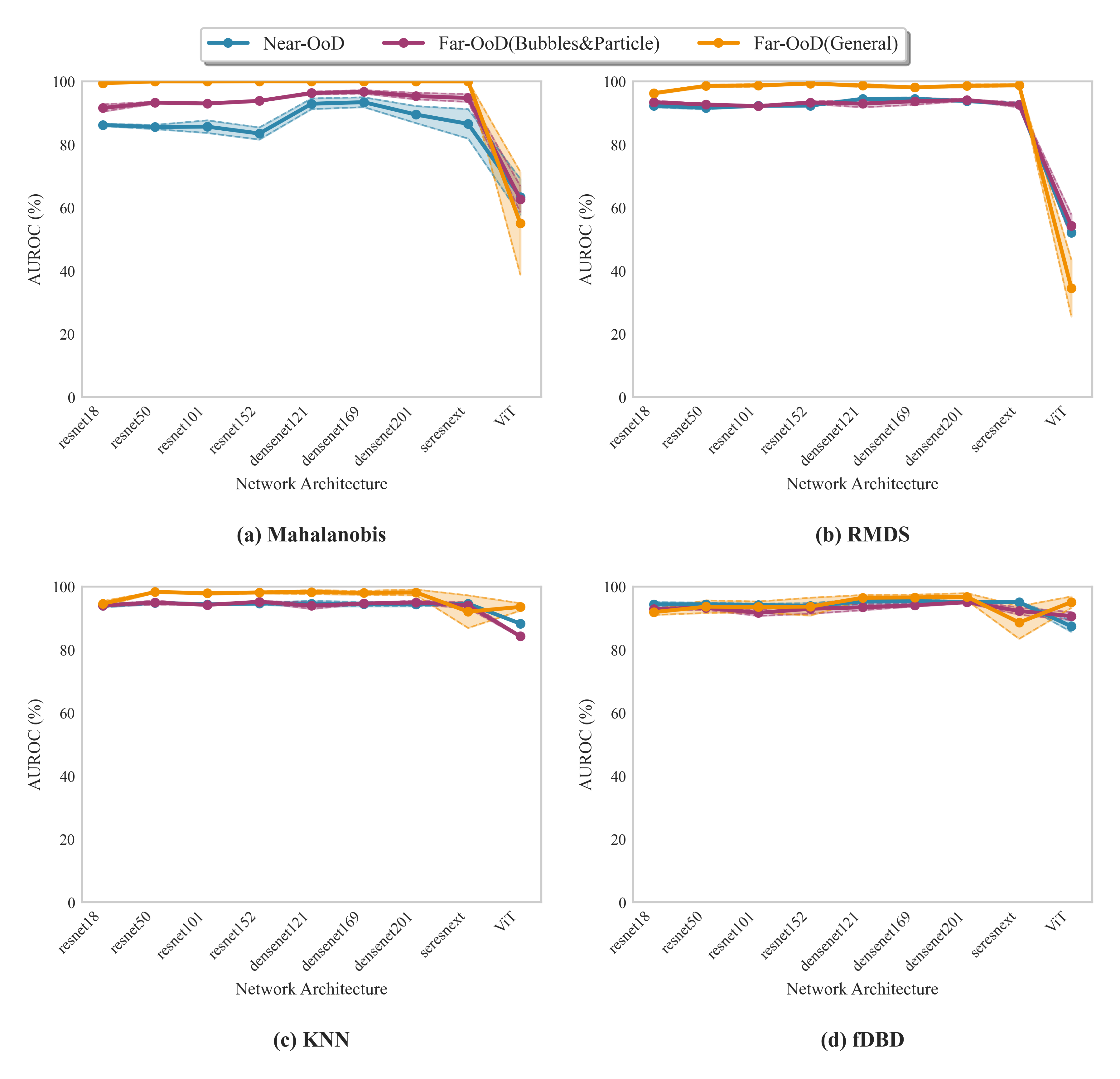}
    \caption{Distance-based Methods. The solid points on the line graph represent the average values, with the standard deviation range illustrated by the shaded area between the dashed lines.}
    \label{fig:Distance-based Methods}
\end{figure*}

\begin{figure*}
    \centering
    \includegraphics[width=1\linewidth]{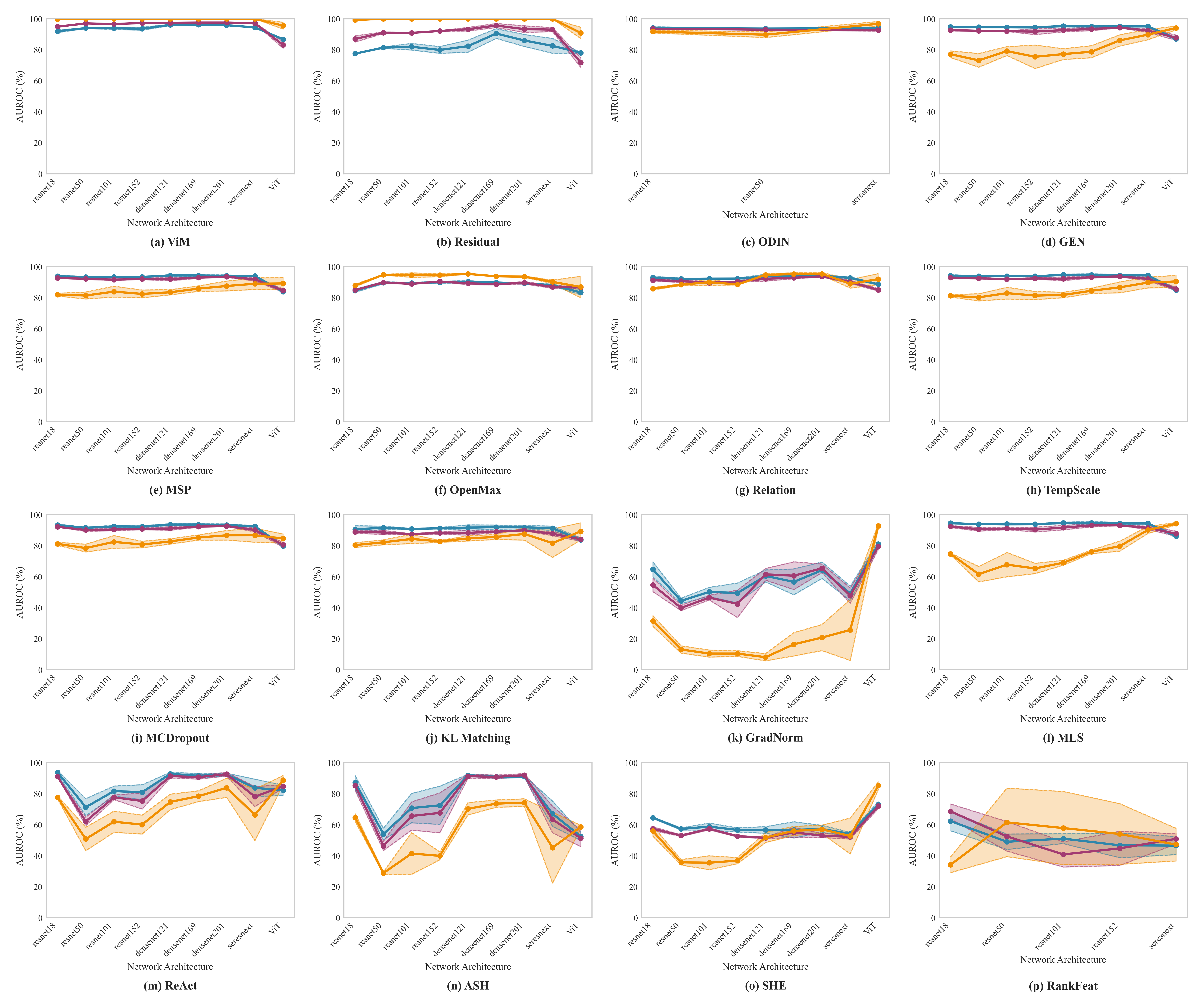}
    \caption{Classification-based Methods. The solid points on the line graph represent the average values, with the standard deviation range illustrated by the shaded area between the dashed lines.}
    \label{fig:Classification-based Methods}
\end{figure*}

\begin{figure*}[!htbp]
    \centering
    \includegraphics[width=1\linewidth]{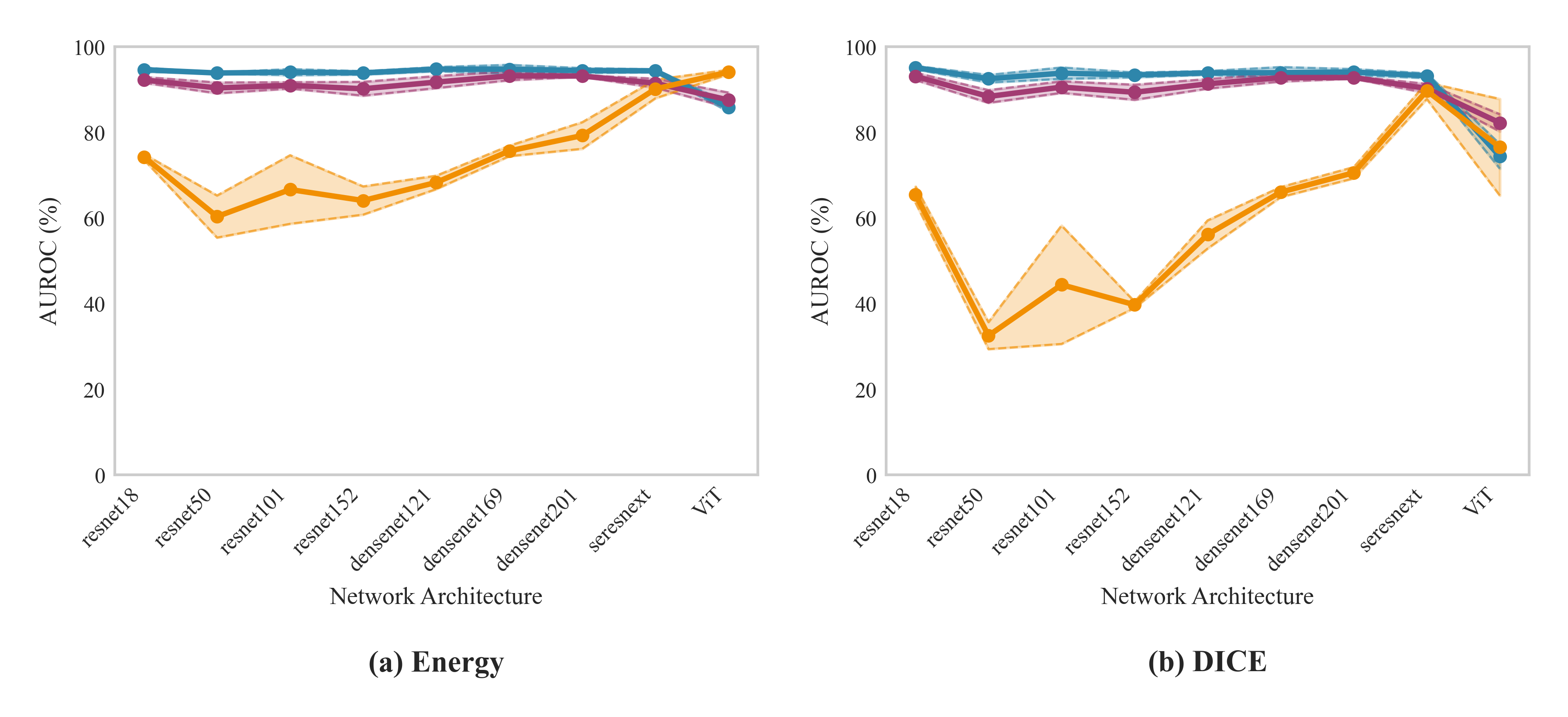}
    \caption{Density-based Methods. The solid points on the line graph represent the average values, with the standard deviation range illustrated by the shaded area between the dashed lines.}
    \label{fig:Density-based Methods}
\end{figure*}

To investigate the influence of different network architectures on OoD detection performance, we designed and conducted an ablation study where we only replaced the network backbone models. Each network was trained three times using different random seeds, and we report the mean and standard deviation of their AUROC values on the Near-OoD, Far-OoD (Bubbles \& Particles), and Far-OoD (General) datasets. For methods requiring hyperparameter tuning, we performed an extensive search for each backbone to ensure the best performance is reported. The experimental results are shown in \cref{fig:Distance-based Methods,fig:Classification-based Methods,fig:Density-based Methods}. We observed that some methods, such as GradNorm, ReAct, ASH, and SHE, exhibit strong dependence on the underlying network, while others, including KNN, fDBD, Relation, and ViM, are less sensitive. This highlights the importance of considering the chosen network architecture when evaluating OoD detection results.

\subsection{A Good Closed-set Classifier Is All You Need?}
% 为了探究OoD表现与分类器准确率的关系，我们选择了五种方法，包括MSP基线、ViM、Energy、KNN和Mahalanobis，并选择了4种常用的网络架构：ResNet-18、ResNet-50、DenseNet-121和ViT，在Near-OoD、Far-OoD(Bubbles & Particles)和Far-OoD(General) benchmark上进行测试。实验严格遵循OpenOoD Guidelines。结果给出了AUROC(OoD 方法表现)与分类精度(ID 分类准确率)的对比。我们发现，对于Near-OoD和Far-OoD(Bubbles & Particles)数据，闭集精度与OoD表现存在正相关关系。对于Near-OoD，ACC与AUROC之间的 Spearman ρ =0.667，p=0.000；对于Far-OoD(Bubbles & Particles)，ACC与AUROC之间的 Spearman ρ =0.609，p=0.004，说明均显著。两个指标之间存在线性相关关系。但对于Far-OoD(general)，ACC与AUROC之间的 Spearman ρ =0.2482，p=0.2914，并不显著。我们认为Near-OoD、Far-OoD(Bubbles & Particles)数据更多的是语义偏移，因此更强的clsose-set分类效果，在这两个OoD基准上的表现可能会更好。但是如果对于非常偏移的数据集(我们的far-ood(general))，那么这样的观察可能就不会生效了。

To investigate the relationship between OoD detection performance and classifier accuracy, we selected five representative methods: MSP, ViM, Energy, KNN, and Mahalanobis. We evaluated them across four common network architectures—ResNet-18, ResNet-50, DenseNet-121, and ViT—on our Near-OoD, Far-OoD (Bubbles \& Particles), and Far-OoD (General) benchmarks, strictly following the OpenOoD guidelines \cite{zhang2023openood}.

\begin{figure*}[!htbp]
    \centering
    \includegraphics[width=1\linewidth]{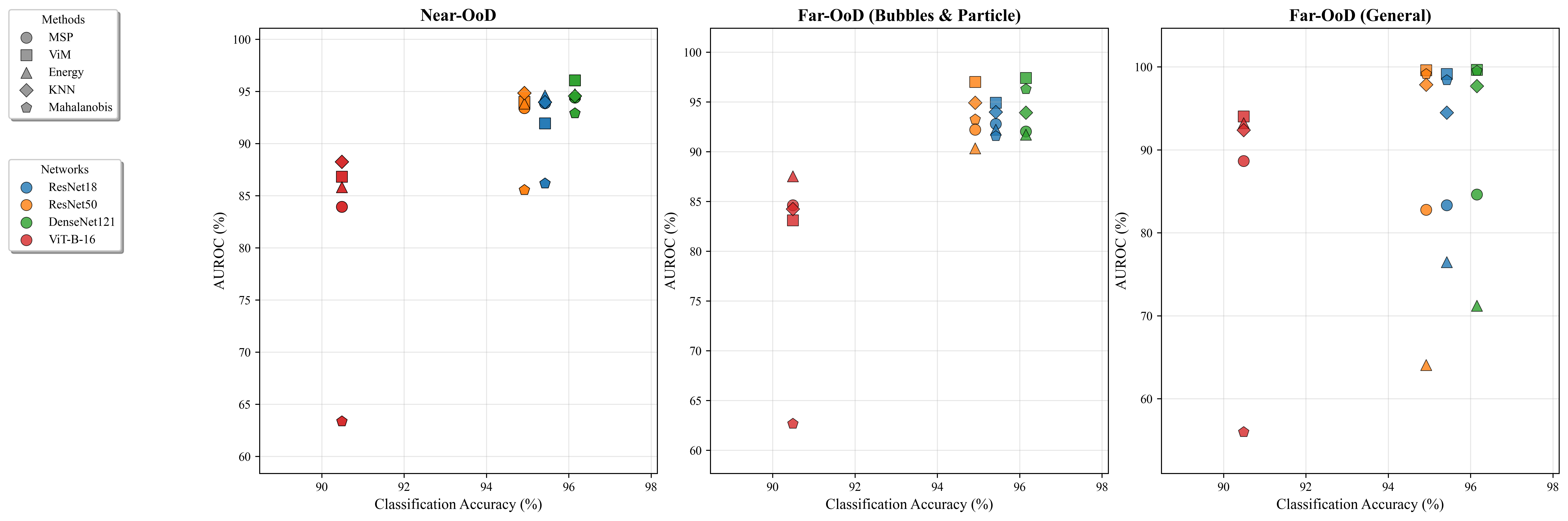}
    \caption{Correlation Between ID Classification Accuracy and OoD Detection Performance. We selected five representative methods: MSP, ViM, Energy, KNN, and Mahalanobis, then we evaluated these methods using four common network architectures: ResNet-18, ResNet-50, DenseNet-121, and ViT, on our Near-OoD, Far-OoD (Bubbles \& Particles), and Far-OoD (General) benchmarks. The average performance of these methods across different architectures was plotted on scatter graphs to visually analyze their correlation.}
    \label{fig:scatter}
\end{figure*}

\Cref{fig:scatter} reveals a significant positive correlation between closed-set classification accuracy (ACC) and OoD detection performance (AUROC) for OoD data with semantic shifts. Specifically, for Near-OoD, the Spearman's $\rho$ correlation coefficient was 0.667 (p $<$ 0.001); for Far-OoD (Bubbles \& Particles), it was 0.609 (p $<$ 0.005), both of which are statistically significant. This suggests that for data with moderate semantic shifts, a stronger classifier generally learns more discriminative feature representations, which in turn improves OoD detection \cite{vaze2021open}. However, for the semantically disjoint Far-OoD (General) data, we observed no significant correlation between ACC and AUROC (Spearman's $\rho$ = 0.248, p = 0.291). This indicates that when OoD samples are highly dissimilar to the ID distribution, simply improving the closed-set classifier's performance is not a sufficient guarantee for better OoD detection.

\section{Network Results}
\label{nerwork result}

\subsection{ResNet-18}
\Cref{18-far,18-near} show the comprehensive performance of the ResNet-18 network on the Far-OoD and Near-OoD benchmarks.

\begin{table*}[htbp!]
    \centering
    \scriptsize
\begin{tabularx}{\textwidth}{@{}l *{10}{Y}@{}} % 使用tabularx，总宽度为\textwidth，Y列会自动调整宽度
    \toprule % 顶部粗线
    \multirow{2}{*}{\parbox[c]{1.5cm}{%
        \centering
        \vspace{3ex}% 在“Method”文本的上方增加 0.5ex 的空间
        \textbf{Method}%
    }} & \multicolumn{5}{c}{\textbf{Far-OoD(Bubbles \& Particles)}} & \multicolumn{5}{c}{\textbf{Far-OoD(General)}} \\
    \cmidrule(lr){2-6} \cmidrule(lr){7-11} % 部分横线
    & \textbf{FPR95-ID}↓ & \textbf{FPR95-OoD}↓ & \textbf{FPR99-ID}↓ & \textbf{FPR99-OoD}↓ & \textbf{AUROC}↑ & \textbf{FPR95-ID}↓ & \textbf{FPR95-OoD}↓ & \textbf{FPR99-ID}↓ & \textbf{FPR99-OoD}↓ & \textbf{AUROC}↑ \\
    \midrule % 中间细线
        ASH & 58.06\tiny $\pm$ 13.75 & 60.48 \tiny $\pm$ 4.63 & 76.16 \tiny $\pm$ 9.10 & 83.78 \tiny $\pm$ 2.39 & 85.23 \tiny $\pm$ 3.29 & 80.65 \tiny $\pm$ 6.34 & 88.07 \tiny $\pm$ 1.87 & 91.36 \tiny $\pm$ 2.56 & 89.47 \tiny $\pm$ 2.35 & 64.49 \tiny $\pm$ 2.54 \\ 
        DICE & 33.79 \tiny $\pm$ 3.18 & 30.72 \tiny $\pm$ 3.19 & 64.11 \tiny $\pm$ 6.26 & 65.58 \tiny $\pm$ 3.35 & 93.05 \tiny $\pm$ 0.85 & 74.18 \tiny $\pm$ 3.78 & 85.83 \tiny $\pm$ 1.14 & 89.12 \tiny $\pm$ 4.23 & 87.60 \tiny $\pm$ 1.12 & 65.47 \tiny $\pm$ 1.90 \\ 
        MCDropout & 42.77 \tiny $\pm$ 1.26 & 29.35 \tiny $\pm$ 0.36 & 75.45 \tiny $\pm$ 2.12 & 62.34 \tiny $\pm$ 2.16 & 92.16 \tiny $\pm$ 0.20 & 64.11 \tiny $\pm$ 3.35 & 76.76 \tiny $\pm$ 4.80 & 89.41 \tiny $\pm$ 1.73 & 89.66 \tiny $\pm$ 2.54 & 81.17 \tiny $\pm$ 1.01 \\ 
        Energy & 37.64 \tiny $\pm$ 3.13 & 31.83 \tiny $\pm$ 2.44 & 72.61 \tiny $\pm$ 2.31 & 74.35 \tiny $\pm$ 3.35 & 92.22 \tiny $\pm$ 0.68 & 64.88 \tiny $\pm$ 3.86 & 84.83 \tiny $\pm$ 1.29 & 86.99 \tiny $\pm$ 2.98 & 88.29 \tiny $\pm$ 1.38 & 74.24 \tiny $\pm$ 0.66 \\ 
        fDBD & 36.18 \tiny $\pm$ 1.68 & 30.42 \tiny $\pm$ 1.67 & 73.00 \tiny $\pm$ 4.39 & 58.96 \tiny $\pm$ 4.60 & 92.91 \tiny $\pm$ 0.37 & 40.78 \tiny $\pm$ 3.29 & 33.36 \tiny $\pm$ 4.00 & 75.89 \tiny $\pm$ 1.66 & 57.54 \tiny $\pm$ 6.60 & 91.89 \tiny $\pm$ 0.89 \\
\hline
    \end{tabularx}
    \caption{Far-OoD on ResNet-18.}
    \label{18-far}
\end{table*}

\begin{table*}[!htbp]
    \centering
    \scriptsize
    
    \begin{tabularx}{\textwidth}{l >{\Centering\arraybackslash}X >{\Centering\arraybackslash}p{2.5cm} >{\Centering\arraybackslash}X >{\Centering\arraybackslash}p{2.5cm} >{\Centering\arraybackslash}X} 
    \toprule % 顶部粗线
        \textbf{Method} & \textbf{FPR95-ID}↓ & \textbf{FPR95-OoD}↓ & \textbf{FPR99-ID}↓ & \textbf{FPR99-OoD}↓ & \textbf{AUROC}↑ \\ 
        \midrule % 中间细线
        
        ASH & 52.36 \tiny $\pm$ 16.96 & 53.45 \tiny $\pm$ 11.25 & 70.22 \tiny $\pm$ 11.39 & 82.71 \tiny $\pm$ 7.61 & 87.14 \tiny $\pm$ 4.49 \\ 
        DICE & 26.89 \tiny $\pm$ 3.29 & 19.02 \tiny $\pm$ 1.78 & 58.48 \tiny $\pm$ 1.47 & 54.73 \tiny $\pm$ 7.30 & 95.09 \tiny $\pm$ 0.40 \\ 
        MCDropout & 40.79 \tiny $\pm$ 2.50 & 24.47 \tiny $\pm$ 2.65 & 73.31 \tiny $\pm$ 1.28 & 46.51 \tiny $\pm$ 8.95 & 93.26 \tiny $\pm$ 0.79 \\ 
        Energy & 28.82 \tiny $\pm$ 3.17 & 20.56 \tiny $\pm$ 1.16 & 65.38 \tiny $\pm$ 2.78 & 56.55 \tiny $\pm$ 3.85 & 94.60 \tiny $\pm$ 0.48 \\ 
        fDBD & 34.24 \tiny $\pm$ 1.62 & 21.29 \tiny $\pm$ 2.46 & 71.24 \tiny $\pm$ 2.13 & 35.39 \tiny $\pm$ 5.12 & 94.37 \tiny $\pm$ 0.70 \\ 
        GEN & 29.08 \tiny $\pm$ 3.58 & 20.06 \tiny $\pm$ 1.47 & 64.74 \tiny $\pm$ 2.83 & 47.65 \tiny $\pm$ 9.13 & 94.72 \tiny $\pm$ 0.51 \\ 
        GradNorm & 79.11 \tiny $\pm$ 11.18 & 88.15 \tiny $\pm$ 2.82 & 88.05 \tiny $\pm$ 9.47 & 97.37 \tiny $\pm$ 1.19 & 64.77 \tiny $\pm$ 4.85 \\
\end{tabularx}
    \caption{Near-OoD on ResNet-18.}
    \label{18-near}
\end{table*}

\subsection{ResNet-50}
\Cref{50-far,50-near} show the comprehensive performance of the ResNet-50 network on the Far-OoD and Near-OoD benchmarks.

\begin{table*}[htbp!]
    \centering
    \scriptsize
    \begin{tabularx}{\textwidth}{@{}l *{10}{Y}@{}} % 使用tabularx，总宽度为\textwidth，Y列会自动调整宽度
    \toprule % 顶部粗线
    \multirow{2}{*}{\parbox[c]{1.5cm}{%
        \centering
        \vspace{3ex}% 在“Method”文本的上方增加 0.5ex 的空间
        \textbf{Method}%
    }} & \multicolumn{5}{c}{\textbf{Far-OoD(Bubbles \& Particles)}} & \multicolumn{5}{c}{\textbf{Far-OoD(General)}} \\
    \cmidrule(lr){2-6} \cmidrule(lr){7-11} % 部分横线
    & \textbf{FPR95-ID}↓ & \textbf{FPR95-OoD}↓ & \textbf{FPR99-ID}↓ & \textbf{FPR99-OoD}↓ & \textbf{AUROC}↑ & \textbf{FPR95-ID}↓ & \textbf{FPR95-OoD}↓ & \textbf{FPR99-ID}↓ & \textbf{FPR99-OoD}↓ & \textbf{AUROC}↑ \\
    \midrule % 中间细线
        ASH & 99.97 \tiny $\pm$ 0.03 & 90.10 \tiny $\pm$ 2.23 & 100.00 \tiny $\pm$ 0.00 & 98.45 \tiny $\pm$ 0.57 & 46.26 \tiny $\pm$ 3.00 & 99.99 \tiny $\pm$ 0.01 & 98.02 \tiny $\pm$ 0.79 & 100.00 \tiny $\pm$ 0.00 & 99.25 \tiny $\pm$ 0.53 & 28.73 \tiny $\pm$ 0.78 \\ 
        DICE & 42.40 \tiny $\pm$ 3.66 & 54.83 \tiny $\pm$ 6.83 & 66.92 \tiny $\pm$ 4.10 & 83.61 \tiny $\pm$ 4.51 & 88.35 \tiny $\pm$ 1.49 & 97.45 \tiny $\pm$ 2.36 & 96.69 \tiny $\pm$ 2.52 & 99.72 \tiny $\pm$ 0.12 & 98.39 \tiny $\pm$ 1.73 & 32.51 \tiny $\pm$ 3.16 \\ 
        MCDropout & 51.32 \tiny $\pm$ 3.88 & 38.16 \tiny $\pm$ 3.53 & 80.02 \tiny $\pm$ 1.36 & 71.35 \tiny $\pm$ 5.67 & 90.11 \tiny $\pm$ 0.88 & 69.19 \tiny $\pm$ 4.91 & 82.21 \tiny $\pm$ 3.82 & 91.96 \tiny $\pm$ 1.66 & 91.39 \tiny $\pm$ 1.14 & 78.46 \tiny $\pm$ 2.51 \\ 
        Energy & 39.93 \tiny $\pm$ 2.84 & 46.27 \tiny $\pm$ 8.14 & 70.97 \tiny $\pm$ 3.80 & 83.07 \tiny $\pm$ 4.01 & 90.35 \tiny $\pm$ 1.27 & 84.47 \tiny $\pm$ 5.14 & 90.66 \tiny $\pm$ 2.77 & 98.35 \tiny $\pm$ 0.35 & 95.01 \tiny $\pm$ 3.43 & 60.31 \tiny $\pm$ 4.92 \\ 
        fDBD & 35.51 \tiny $\pm$ 4.02 & 27.46 \tiny $\pm$ 2.78 & 72.05 \tiny $\pm$ 1.68 & 54.64 \tiny $\pm$ 5.32 & 93.28 \tiny $\pm$ 0.68 & 31.00 \tiny $\pm$ 10.40 & 27.69 \tiny $\pm$ 6.56 & 67.48 \tiny $\pm$ 9.32 & 56.01 \tiny $\pm$ 13.61 & 93.62 \tiny $\pm$ 2.00 \\
\hline
    \end{tabularx}
    \caption{Far-OoD on ResNet-50.}
    \label{50-far}
\end{table*}

\begin{table*}[!htbp]
    \centering
    \scriptsize
    
    \begin{tabularx}{\textwidth}{l >{\Centering\arraybackslash}X >{\Centering\arraybackslash}p{2.5cm} >{\Centering\arraybackslash}X >{\Centering\arraybackslash}p{2.5cm} >{\Centering\arraybackslash}X} 
    \toprule % 顶部粗线
        \textbf{Method} & \textbf{FPR95-ID}↓ & \textbf{FPR95-OoD}↓ & \textbf{FPR99-ID}↓ & \textbf{FPR99-OoD}↓ & \textbf{AUROC}↑ \\ 
        \midrule % 中间细线
        ASH & 99.97 \tiny $\pm$ 0.04 & 79.90 \tiny $\pm$ 1.24 & 100.00 \tiny $\pm$ 0.00 & 92.13 \tiny $\pm$ 0.27 & 53.95 \tiny $\pm$ 3.71 \\ 
        DICE & 31.85 \tiny $\pm$ 3.57 & 38.15 \tiny $\pm$ 4.44 & 58.01 \tiny $\pm$ 3.47 & 70.70 \tiny $\pm$ 6.06 & 92.49 \tiny $\pm$ 0.89 \\ 
        MCDropout & 50.50 \tiny $\pm$ 0.25 & 30.25 \tiny $\pm$ 1.12 & 80.36 \tiny $\pm$ 1.90 & 50.44 \tiny $\pm$ 3.78 & 91.56 \tiny $\pm$ 0.22 \\ 
        Energy & 31.59 \tiny $\pm$ 1.18 & 25.66 \tiny $\pm$ 0.80 & 67.42 \tiny $\pm$ 2.50 & 59.28 \tiny $\pm$ 5.49 & 93.83 \tiny $\pm$ 0.15 \\ 
        fDBD & 33.57 \tiny $\pm$ 3.83 & 22.00 \tiny $\pm$ 1.78 & 72.61 \tiny $\pm$ 3.74 & 35.61 \tiny $\pm$ 1.17 & 94.39 \tiny $\pm$ 0.54 \\ 
        GEN & 30.19 \tiny $\pm$ 1.60 & 20.49 \tiny $\pm$ 2.33 & 67.77 \tiny $\pm$ 1.79 & 41.95 \tiny $\pm$ 5.76 & 94.62 \tiny $\pm$ 0.41 \\ 
        GradNorm & 100.00 \tiny $\pm$ 0.00 & 93.15 \tiny $\pm$ 2.66 & 100.00 \tiny $\pm$ 0.00 & 98.10 \tiny $\pm$ 0.44 & 44.39 \tiny $\pm$ 1.73 \\
\end{tabularx}
    \caption{Near-OoD on ResNet-50.}
    \label{50-near}
\end{table*}

\subsection{ResNet-101}
\Cref{101-far,101-near} show the comprehensive performance of the ResNet-101 network on the Far-OoD and Near-OoD benchmarks.

\begin{table*}[htbp!]
    \centering
    \scriptsize
    \begin{tabularx}{\textwidth}{@{}l *{10}{Y}@{}} % 使用tabularx，总宽度为\textwidth，Y列会自动调整宽度
    \toprule % 顶部粗线
    \multirow{2}{*}{\parbox[c]{1.5cm}{%
        \centering
        \vspace{3ex}% 在“Method”文本的上方增加 0.5ex 的空间
        \textbf{Method}%
    }} & \multicolumn{5}{c}{\textbf{Far-OoD(Bubbles \& Particles)}} & \multicolumn{5}{c}{\textbf{Far-OoD(General)}} \\
    \cmidrule(lr){2-6} \cmidrule(lr){7-11} % 部分横线
    & \textbf{FPR95-ID}↓ & \textbf{FPR95-OoD}↓ & \textbf{FPR99-ID}↓ & \textbf{FPR99-OoD}↓ & \textbf{AUROC}↑ & \textbf{FPR95-ID}↓ & \textbf{FPR95-OoD}↓ & \textbf{FPR99-ID}↓ & \textbf{FPR99-OoD}↓ & \textbf{AUROC}↑ \\
    \midrule % 中间细线
        ASH & 89.21 \tiny $\pm$ 10.26 & 80.84 \tiny $\pm$ 7.03 & 97.80 \tiny $\pm$ 2.68 & 94.74 \tiny $\pm$ 3.25 & 65.58 \tiny $\pm$ 9.12 & 98.02 \tiny $\pm$ 2.72 & 94.03 \tiny $\pm$ 3.01 & 99.92 \tiny $\pm$ 0.11 & 97.72 \tiny $\pm$ 1.10 & 41.36 \tiny $\pm$ 13.47 \\ 
        DICE & 35.23 \tiny $\pm$ 1.81 & 49.27 \tiny $\pm$ 8.09 & 61.51 \tiny $\pm$ 1.68 & 79.66 \tiny $\pm$ 5.34 & 90.54 \tiny $\pm$ 1.36 & 90.30 \tiny $\pm$ 5.65 & 91.33 \tiny $\pm$ 4.34 & 99.14 \tiny $\pm$ 0.34 & 94.77 \tiny $\pm$ 3.98 & 44.39 \tiny $\pm$ 13.83 \\ 
        MCDropout & 49.91 \tiny $\pm$ 2.62 & 36.74 \tiny $\pm$ 2.26 & 79.26 \tiny $\pm$ 1.10 & 67.59 \tiny $\pm$ 6.06 & 90.43 \tiny $\pm$ 0.72 & 61.17 \tiny $\pm$ 7.89 & 74.36 \tiny $\pm$ 9.57 & 89.11 \tiny $\pm$ 2.82 & 88.64 \tiny $\pm$ 2.54 & 82.45 \tiny $\pm$ 4.07 \\ 
        Energy & 37.85 \tiny $\pm$ 1.79 & 43.57 \tiny $\pm$ 4.56 & 70.31 \tiny $\pm$ 1.26 & 82.03 \tiny $\pm$ 2.65 & 90.94 \tiny $\pm$ 0.75 & 76.22 \tiny $\pm$ 9.19 & 86.68 \tiny $\pm$ 3.81 & 97.61 \tiny $\pm$ 1.37 & 91.26 \tiny $\pm$ 2.86 & 66.62 \tiny $\pm$ 8.01 \\ 
        fDBD & 41.97 \tiny $\pm$ 1.81 & 33.48 \tiny $\pm$ 4.28 & 75.91 \tiny $\pm$ 3.06 & 61.57 \tiny $\pm$ 6.38 & 91.65 \tiny $\pm$ 0.95 & 30.61 \tiny $\pm$ 6.99 & 27.74 \tiny $\pm$ 7.29 & 71.34 \tiny $\pm$ 8.95 & 58.91 \tiny $\pm$ 12.42 & 93.53 \tiny $\pm$ 1.74 \\
\hline
    \end{tabularx}
    \caption{Far-OoD on ResNet-101.}
    \label{101-far}
\end{table*}

\begin{table*}[!htbp]
    \centering
    \scriptsize
    
    \begin{tabularx}{\textwidth}{l >{\Centering\arraybackslash}X >{\Centering\arraybackslash}p{2.5cm} >{\Centering\arraybackslash}X >{\Centering\arraybackslash}p{2.5cm} >{\Centering\arraybackslash}X} 
    \toprule % 顶部粗线
        \textbf{Method} & \textbf{FPR95-ID}↓ & \textbf{FPR95-OoD}↓ & \textbf{FPR99-ID}↓ & \textbf{FPR99-OoD}↓ & \textbf{AUROC}↑ \\ 
        \midrule % 中间细线
        ASH & 86.22 \tiny $\pm$ 14.10 & 69.81 \tiny $\pm$ 9.94 & 96.91 \tiny $\pm$ 4.24 & 90.58 \tiny $\pm$ 3.46 & 70.67 \tiny $\pm$ 9.49 \\ 
        DICE & 26.34 \tiny $\pm$ 4.08 & 31.27 \tiny $\pm$ 8.87 & 57.60 \tiny $\pm$ 2.81 & 64.18 \tiny $\pm$ 12.15 & 93.80 \tiny $\pm$ 1.33 \\ 
        MCDropout & 45.54 \tiny $\pm$ 2.93 & 26.60 \tiny $\pm$ 2.63 & 76.52 \tiny $\pm$ 1.11 & 49.55 \tiny $\pm$ 4.40 & 92.43 \tiny $\pm$ 0.83 \\ 
        Energy & 30.16 \tiny $\pm$ 1.92 & 24.88 \tiny $\pm$ 4.18 & 67.10 \tiny $\pm$ 3.08 & 56.61 \tiny $\pm$ 8.82 & 94.03 \tiny $\pm$ 0.71 \\ 
        fDBD & 35.31 \tiny $\pm$ 0.50 & 22.60 \tiny $\pm$ 1.51 & 70.75 \tiny $\pm$ 2.99 & 37.86 \tiny $\pm$ 5.20 & 94.15 \tiny $\pm$ 0.44 \\ 
        GEN & 32.52 \tiny $\pm$ 2.61 & 20.78 \tiny $\pm$ 1.64 & 67.02 \tiny $\pm$ 2.97 & 42.77 \tiny $\pm$ 1.12 & 94.55 \tiny $\pm$ 0.38 \\ 
        GradNorm & 98.60 \tiny $\pm$ 0.93 & 91.76 \tiny $\pm$ 0.26 & 99.65 \tiny $\pm$ 0.26 & 98.67 \tiny $\pm$ 0.04 & 50.19 \tiny $\pm$ 2.93 \\
\end{tabularx}
    \caption{Near-OoD on ResNet-101.}
    \label{101-near}
\end{table*}

\subsection{ResNet-152}
\Cref{152-far,152-near} show the comprehensive performance of the ResNet-152 network on the Far-OoD and Near-OoD benchmarks.

\begin{table*}[htbp!]
    \centering
    \scriptsize
    \begin{tabularx}{\textwidth}{@{}l *{10}{Y}@{}} % 使用tabularx，总宽度为\textwidth，Y列会自动调整宽度
    \toprule % 顶部粗线
    \multirow{2}{*}{\parbox[c]{1.5cm}{%
        \centering
        \vspace{3ex}% 在“Method”文本的上方增加 0.5ex 的空间
        \textbf{Method}%
    }} & \multicolumn{5}{c}{\textbf{Far-OoD(Bubbles \& Particles)}} & \multicolumn{5}{c}{\textbf{Far-OoD(General)}} \\
    \cmidrule(lr){2-6} \cmidrule(lr){7-11} % 部分横线
    & \textbf{FPR95-ID}↓ & \textbf{FPR95-OoD}↓ & \textbf{FPR99-ID}↓ & \textbf{FPR99-OoD}↓ & \textbf{AUROC}↑ & \textbf{FPR95-ID}↓ & \textbf{FPR95-OoD}↓ & \textbf{FPR99-ID}↓ & \textbf{FPR99-OoD}↓ & \textbf{AUROC}↑ \\
    \midrule % 中间细线
        ASH & 81.97 \tiny $\pm$ 16.72 & 79.05 \tiny $\pm$ 7.56 & 94.38 \tiny $\pm$ 7.11 & 92.79 \tiny $\pm$ 4.46 & 67.57 \tiny $\pm$ 12.88 & 97.93 \tiny $\pm$ 2.26 & 93.16 \tiny $\pm$ 2.20 & 99.93 \tiny $\pm$ 0.09 & 96.07 \tiny $\pm$ 1.85 & 39.85 \tiny $\pm$ 2.55 \\ 
        DICE & 38.92 \tiny $\pm$ 1.97 & 52.11 \tiny $\pm$ 10.44 & 65.66 \tiny $\pm$ 0.48 & 81.71 \tiny $\pm$ 7.39 & 89.33 \tiny $\pm$ 1.77 & 92.27 \tiny $\pm$ 2.08 & 90.51 \tiny $\pm$ 1.52 & 99.17 \tiny $\pm$ 0.67 & 93.41 \tiny $\pm$ 1.65 & 39.71 \tiny $\pm$ 0.74 \\ 
        MCDropout & 49.36 \tiny $\pm$ 1.53 & 33.50 \tiny $\pm$ 2.28 & 79.55 \tiny $\pm$ 1.15 & 63.89 \tiny $\pm$ 3.01 & 90.84 \tiny $\pm$ 0.53 & 65.04 \tiny $\pm$ 2.76 & 77.92 \tiny $\pm$ 6.53 & 91.18 \tiny $\pm$ 0.93 & 89.36 \tiny $\pm$ 1.07 & 80.74 \tiny $\pm$ 2.11 \\ 
        Energy & 41.64 \tiny $\pm$ 2.03 & 47.06 \tiny $\pm$ 14.22 & 73.47 \tiny $\pm$ 3.26 & 83.51 \tiny $\pm$ 7.05 & 90.15 \tiny $\pm$ 1.62 & 80.56 \tiny $\pm$ 4.20 & 87.08 \tiny $\pm$ 0.60 & 98.02 \tiny $\pm$ 0.91 & 89.59 \tiny $\pm$ 1.29 & 64.05 \tiny $\pm$ 3.30 \\ 
        fDBD & 38.52 \tiny $\pm$ 6.57 & 27.61 \tiny $\pm$ 5.63 & 74.17 \tiny $\pm$ 5.66 & 51.27 \tiny $\pm$ 11.03 & 92.97 \tiny $\pm$ 1.59 & 31.02 \tiny $\pm$ 12.07 & 26.73 \tiny $\pm$ 11.40 & 68.82 \tiny $\pm$ 16.81 & 50.92 \tiny $\pm$ 16.62 & 93.64 \tiny $\pm$ 2.83 \\
\hline
    \end{tabularx}
    \caption{Far-OoD on ResNet-152.}
    \label{152-far}
\end{table*}

\begin{table*}[!htbp]
    \centering
    \scriptsize
    
    \begin{tabularx}{\textwidth}{l >{\Centering\arraybackslash}X >{\Centering\arraybackslash}p{2.5cm} >{\Centering\arraybackslash}X >{\Centering\arraybackslash}p{2.5cm} >{\Centering\arraybackslash}X} 
    \toprule % 顶部粗线
        \textbf{Method} & \textbf{FPR95-ID}↓ & \textbf{FPR95-OoD}↓ & \textbf{FPR99-ID}↓ & \textbf{FPR99-OoD}↓ & \textbf{AUROC}↑ \\ 
        \midrule % 中间细线
        ASH & 79.05 \tiny $\pm$ 18.09 & 69.73 \tiny $\pm$ 7.88 & 93.59 \tiny $\pm$ 7.75 & 87.20 \tiny $\pm$ 3.40 & 72.47 \tiny $\pm$ 12.33 \\ 
        DICE & 29.69 \tiny $\pm$ 0.78 & 32.43 \tiny $\pm$ 3.43 & 63.90 \tiny $\pm$ 1.73 & 65.63 \tiny $\pm$ 7.18 & 93.38 \tiny $\pm$ 0.57 \\ 
        MCDropout & 46.18 \tiny $\pm$ 2.84 & 26.57 \tiny $\pm$ 2.13 & 76.60 \tiny $\pm$ 3.25 & 52.30 \tiny $\pm$ 8.33 & 92.30 \tiny $\pm$ 0.61 \\ 
        Energy & 34.44 \tiny $\pm$ 2.49 & 23.60 \tiny $\pm$ 2.37 & 69.58 \tiny $\pm$ 2.83 & 59.68 \tiny $\pm$ 6.45 & 93.86 \tiny $\pm$ 0.40 \\ 
        fDBD & 35.34 \tiny $\pm$ 5.52 & 24.20 \tiny $\pm$ 2.61 & 73.16 \tiny $\pm$ 5.03 & 40.70 \tiny $\pm$ 0.91 & 93.88 \tiny $\pm$ 0.87 \\ 
        GEN & 33.07 \tiny $\pm$ 3.56 & 20.14 \tiny $\pm$ 0.64 & 69.99 \tiny $\pm$ 4.35 & 45.10 \tiny $\pm$ 4.77 & 94.46 \tiny $\pm$ 0.38 \\ 
        GradNorm & 96.77 \tiny $\pm$ 2.75 & 92.18 \tiny $\pm$ 1.11 & 99.04 \tiny $\pm$ 0.72 & 97.15 \tiny $\pm$ 0.80 & 49.45 \tiny $\pm$ 6.45 \\
\end{tabularx}
    \caption{Near-OoD on ResNet-152.}
    \label{152-near}
\end{table*}

\subsection{DenseNet-121}
\Cref{121-far,121-near} show the comprehensive performance of the DenseNet-121 network on the Far-OoD and Near-OoD benchmarks.

\begin{table*}[htbp!]
    \centering
    \scriptsize
    \begin{tabularx}{\textwidth}{@{}l *{10}{Y}@{}} % 使用tabularx，总宽度为\textwidth，Y列会自动调整宽度
    \toprule % 顶部粗线
    \multirow{2}{*}{\parbox[c]{1.5cm}{%
        \centering
        \vspace{3ex}% 在“Method”文本的上方增加 0.5ex 的空间
        \textbf{Method}%
    }} & \multicolumn{5}{c}{\textbf{Far-OoD(Bubbles \& Particles)}} & \multicolumn{5}{c}{\textbf{Far-OoD(General)}} \\
    \cmidrule(lr){2-6} \cmidrule(lr){7-11} % 部分横线
    & \textbf{FPR95-ID}↓ & \textbf{FPR95-OoD}↓ & \textbf{FPR99-ID}↓ & \textbf{FPR99-OoD}↓ & \textbf{AUROC}↑ & \textbf{FPR95-ID}↓ & \textbf{FPR95-OoD}↓ & \textbf{FPR99-ID}↓ & \textbf{FPR99-OoD}↓ & \textbf{AUROC}↑ \\
    \midrule % 中间细线
        ASH & 37.59 \tiny $\pm$ 3.02 & 42.22 \tiny $\pm$ 8.59 & 62.09 \tiny $\pm$ 1.26 & 68.99 \tiny $\pm$ 6.99 & 91.27 \tiny $\pm$ 1.27 & 68.51 \tiny $\pm$ 3.52 & 82.65 \tiny $\pm$ 3.99 & 91.96 \tiny $\pm$ 1.67 & 86.93 \tiny $\pm$ 1.22 & 70.22 \tiny $\pm$ 3.99 \\ 
        DICE & 25.73 \tiny $\pm$ 1.05 & 57.08 \tiny $\pm$ 9.26 & 55.93 \tiny $\pm$ 3.19 & 86.97 \tiny $\pm$ 6.00 & 91.30 \tiny $\pm$ 1.12 & 70.44 \tiny $\pm$ 3.93 & 86.17 \tiny $\pm$ 0.23 & 88.98 \tiny $\pm$ 1.11 & 87.31 \tiny $\pm$ 0.67 & 56.14 \tiny $\pm$ 3.29 \\ 
        MCDropout & 40.09 \tiny $\pm$ 1.28 & 42.52 \tiny $\pm$ 7.55 & 71.91 \tiny $\pm$ 5.09 & 83.35 \tiny $\pm$ 6.45 & 91.09 \tiny $\pm$ 1.13 & 53.58 \tiny $\pm$ 1.88 & 81.29 \tiny $\pm$ 5.01 & 84.37 \tiny $\pm$ 5.49 & 89.56 \tiny $\pm$ 2.26 & 82.76 \tiny $\pm$ 1.72 \\ 
        Energy & 27.66 \tiny $\pm$ 1.39 & 52.45 \tiny $\pm$ 14.86 & 59.70 \tiny $\pm$ 2.85 & 87.10 \tiny $\pm$ 8.43 & 91.71 \tiny $\pm$ 1.42 & 60.98 \tiny $\pm$ 0.65 & 86.13 \tiny $\pm$ 0.72 & 88.87 \tiny $\pm$ 2.99 & 86.98 \tiny $\pm$ 0.61 & 68.28 \tiny $\pm$ 1.57 \\ 
        fDBD & 30.28 \tiny $\pm$ 2.61 & 29.39 \tiny $\pm$ 4.52 & 67.15 \tiny $\pm$ 4.74 & 57.22 \tiny $\pm$ 7.98 & 93.42 \tiny $\pm$ 0.92 & 17.37 \tiny $\pm$ 4.69 & 14.68 \tiny $\pm$ 3.63 & 57.40 \tiny $\pm$ 11.98 & 34.34 \tiny $\pm$ 6.41 & 96.44 \tiny $\pm$ 0.89 \\
\hline
    \end{tabularx}
    \caption{Far-OoD on DenseNet-121.}
    \label{121-far}
\end{table*}

\begin{table*}[!htbp]
    \centering
    \scriptsize
    
    \begin{tabularx}{\textwidth}{l >{\Centering\arraybackslash}X >{\Centering\arraybackslash}p{2.5cm} >{\Centering\arraybackslash}X >{\Centering\arraybackslash}p{2.5cm} >{\Centering\arraybackslash}X} 
    \toprule % 顶部粗线
        \textbf{Method} & \textbf{FPR95-ID}↓ & \textbf{FPR95-OoD}↓ & \textbf{FPR99-ID}↓ & \textbf{FPR99-OoD}↓ & \textbf{AUROC}↑ \\ 
        \midrule % 中间细线
        ASH & 38.23 \tiny $\pm$ 3.10 & 36.06 \tiny $\pm$ 2.86 & 67.45 \tiny $\pm$ 3.41 & 61.35 \tiny $\pm$ 1.62 & 91.86 \tiny $\pm$ 0.69 \\ 
        DICE & 22.17 \tiny $\pm$ 2.63 & 33.61 \tiny $\pm$ 2.68 & 58.19 \tiny $\pm$ 5.58 & 78.94 \tiny $\pm$ 7.13 & 93.86 \tiny $\pm$ 0.43 \\ 
        MCDropout & 36.95 \tiny $\pm$ 5.03 & 24.31 \tiny $\pm$ 2.49 & 69.81 \tiny $\pm$ 7.15 & 57.81 \tiny $\pm$ 1.83 & 93.62 \tiny $\pm$ 0.59 \\ 
        Energy & 23.63 \tiny $\pm$ 3.93 & 21.46 \tiny $\pm$ 2.95 & 57.49 \tiny $\pm$ 4.99 & 73.07 \tiny $\pm$ 10.07 & 94.73 \tiny $\pm$ 0.49 \\ 
        fDBD & 28.06 \tiny $\pm$ 5.33 & 18.78 \tiny $\pm$ 2.67 & 64.04 \tiny $\pm$ 7.54 & 30.93 \tiny $\pm$ 1.18 & 95.29 \tiny $\pm$ 0.77 \\ 
        GEN & 25.44 \tiny $\pm$ 4.35 & 18.11 \tiny $\pm$ 2.26 & 60.78 \tiny $\pm$ 4.84 & 48.69 \tiny $\pm$ 4.52 & 95.33 \tiny $\pm$ 0.47 \\ 
        GradNorm & 80.86 \tiny $\pm$ 3.16 & 90.95 \tiny $\pm$ 0.20 & 86.80 \tiny $\pm$ 1.43 & 97.38 \tiny $\pm$ 0.98 & 60.49 \tiny $\pm$ 3.86 \\
\end{tabularx}
    \caption{Near-OoD on DenseNet-121.}
    \label{121-near}
\end{table*}

\subsection{DenseNet-169}
\Cref{169-far,169-near} show the comprehensive performance of the DenseNet-169 network on the Far-OoD and Near-OoD benchmarks.

\begin{table*}[htbp!]
    \centering
    \scriptsize
    \begin{tabularx}{\textwidth}{@{}l *{10}{Y}@{}} % 使用tabularx，总宽度为\textwidth，Y列会自动调整宽度
    \toprule % 顶部粗线
    \multirow{2}{*}{\parbox[c]{1.5cm}{%
        \centering
        \vspace{3ex}% 在“Method”文本的上方增加 0.5ex 的空间
        \textbf{Method}%
    }} & \multicolumn{5}{c}{\textbf{Far-OoD(Bubbles \& Particles)}} & \multicolumn{5}{c}{\textbf{Far-OoD(General)}} \\
    \cmidrule(lr){2-6} \cmidrule(lr){7-11} % 部分横线
    & \textbf{FPR95-ID}↓ & \textbf{FPR95-OoD}↓ & \textbf{FPR99-ID}↓ & \textbf{FPR99-OoD}↓ & \textbf{AUROC}↑ & \textbf{FPR95-ID}↓ & \textbf{FPR95-OoD}↓ & \textbf{FPR99-ID}↓ & \textbf{FPR99-OoD}↓ & \textbf{AUROC}↑ \\
    \midrule % 中间细线
        ASH & 37.79 \tiny $\pm$ 2.04 & 45.07 \tiny $\pm$ 7.44 & 61.07 \tiny $\pm$ 4.12 & 67.59 \tiny $\pm$ 5.11 & 90.83 \tiny $\pm$ 1.11 & 62.87 \tiny $\pm$ 2.37 & 80.65 \tiny $\pm$ 3.53 & 83.22 \tiny $\pm$ 3.65 & 86.61 \tiny $\pm$ 0.86 & 73.54 \tiny $\pm$ 2.36 \\ 
        DICE & 22.96 \tiny $\pm$ 0.34 & 47.63 \tiny $\pm$ 10.28 & 53.71 \tiny $\pm$ 3.38 & 88.95 \tiny $\pm$ 3.78 & 92.75 \tiny $\pm$ 0.92 & 59.48 \tiny $\pm$ 2.61 & 85.82 \tiny $\pm$ 0.53 & 80.42 \tiny $\pm$ 0.55 & 86.80 \tiny $\pm$ 0.68 & 66.02 \tiny $\pm$ 1.18 \\ 
        MCDropout & 36.42 \tiny $\pm$ 1.68 & 33.33 \tiny $\pm$ 3.18 & 71.47 \tiny $\pm$ 3.09 & 78.33 \tiny $\pm$ 1.50 & 92.31 \tiny $\pm$ 0.40 & 47.48 \tiny $\pm$ 1.64 & 74.19 \tiny $\pm$ 11.14 & 82.36 \tiny $\pm$ 2.14 & 89.87 \tiny $\pm$ 3.67 & 85.27 \tiny $\pm$ 1.68 \\ 
        Energy & 25.28 \tiny $\pm$ 0.79 & 37.72 \tiny $\pm$ 12.26 & 57.56 \tiny $\pm$ 2.51 & 87.72 \tiny $\pm$ 6.04 & 93.16 \tiny $\pm$ 1.02 & 50.71 \tiny $\pm$ 0.82 & 85.49 \tiny $\pm$ 1.15 & 81.47 \tiny $\pm$ 1.73 & 87.53 \tiny $\pm$ 1.20 & 75.63 \tiny $\pm$ 1.23 \\ 
        fDBD & 30.75 \tiny $\pm$ 2.41 & 25.65 \tiny $\pm$ 1.31 & 67.00 \tiny $\pm$ 2.70 & 51.58 \tiny $\pm$ 6.59 & 94.07 \tiny $\pm$ 0.38 & 18.49 \tiny $\pm$ 3.92 & 14.00 \tiny $\pm$ 4.27 & 56.25 \tiny $\pm$ 0.74 & 29.81 \tiny $\pm$ 8.80 & 96.55 \tiny $\pm$ 0.88 \\
\hline
    \end{tabularx}
    \caption{Far-OoD on DenseNet-169.}
    \label{169-far}
\end{table*}

\begin{table*}[!htbp]
    \centering
    \scriptsize
    
    \begin{tabularx}{\textwidth}{l >{\Centering\arraybackslash}X >{\Centering\arraybackslash}p{2.5cm} >{\Centering\arraybackslash}X >{\Centering\arraybackslash}p{2.5cm} >{\Centering\arraybackslash}X} 
    \toprule % 顶部粗线
        \textbf{Method} & \textbf{FPR95-ID}↓ & \textbf{FPR95-OoD}↓ & \textbf{FPR99-ID}↓ & \textbf{FPR99-OoD}↓ & \textbf{AUROC}↑ \\ 
        \midrule % 中间细线
        ASH & 41.03 \tiny $\pm$ 1.21 & 39.30 \tiny $\pm$ 6.88 & 70.31 \tiny $\pm$ 4.63 & 60.17 \tiny $\pm$ 5.14 & 90.85 \tiny $\pm$ 0.97 \\ 
        DICE & 21.79 \tiny $\pm$ 4.54 & 34.73 \tiny $\pm$ 10.01 & 56.35 \tiny $\pm$ 7.85 & 71.57 \tiny $\pm$ 13.31 & 93.91 \tiny $\pm$ 1.31 \\ 
        MCDropout & 35.14 \tiny $\pm$ 3.17 & 24.30 \tiny $\pm$ 3.14 & 71.42 \tiny $\pm$ 3.60 & 61.42 \tiny $\pm$ 11.89 & 93.66 \tiny $\pm$ 0.82 \\ 
        Energy & 22.99 \tiny $\pm$ 4.95 & 24.46 \tiny $\pm$ 4.98 & 57.05 \tiny $\pm$ 5.62 & 65.01 \tiny $\pm$ 16.18 & 94.72 \tiny $\pm$ 1.02 \\ 
        fDBD & 29.95 \tiny $\pm$ 4.24 & 18.18 \tiny $\pm$ 1.43 & 67.25 \tiny $\pm$ 1.04 & 32.54 \tiny $\pm$ 2.52 & 95.36 \tiny $\pm$ 0.57 \\ 
        GEN & 24.16 \tiny $\pm$ 5.43 & 20.35 \tiny $\pm$ 3.40 & 60.81 \tiny $\pm$ 7.15 & 55.39 \tiny $\pm$ 10.77 & 95.10 \tiny $\pm$ 0.85 \\ 
        GradNorm & 80.86 \tiny $\pm$ 6.15 & 92.17 \tiny $\pm$ 3.57 & 88.20 \tiny $\pm$ 3.72 & 97.30 \tiny $\pm$ 0.78 & 56.65 \tiny $\pm$ 8.35 \\
\end{tabularx}
    \caption{Near-OoD on DenseNet-169.}
    \label{169-near}
\end{table*}

\subsection{DenseNet-201}
\Cref{201-far,201-near} show the comprehensive performance of the DenseNet-201 network on the Far-OoD and Near-OoD benchmarks.

\begin{table*}[htbp!]
    \centering
    \scriptsize
    \begin{tabularx}{\textwidth}{@{}l *{10}{Y}@{}} % 使用tabularx，总宽度为\textwidth，Y列会自动调整宽度
    \toprule % 顶部粗线
    \multirow{2}{*}{\parbox[c]{1.5cm}{%
        \centering
        \vspace{3ex}% 在“Method”文本的上方增加 0.5ex 的空间
        \textbf{Method}%
    }} & \multicolumn{5}{c}{\textbf{Far-OoD(Bubbles \& Particles)}} & \multicolumn{5}{c}{\textbf{Far-OoD(General)}} \\
    \cmidrule(lr){2-6} \cmidrule(lr){7-11} % 部分横线
    & \textbf{FPR95-ID}↓ & \textbf{FPR95-OoD}↓ & \textbf{FPR99-ID}↓ & \textbf{FPR99-OoD}↓ & \textbf{AUROC}↑ & \textbf{FPR95-ID}↓ & \textbf{FPR95-OoD}↓ & \textbf{FPR99-ID}↓ & \textbf{FPR99-OoD}↓ & \textbf{AUROC}↑ \\
    \midrule % 中间细线
        ASH & 40.61 \tiny $\pm$ 6.18 & 36.37 \tiny $\pm$ 4.42 & 77.14 \tiny $\pm$ 15.52 & 60.53 \tiny $\pm$ 6.30 & 91.89 \tiny $\pm$ 1.03 & 73.21 \tiny $\pm$ 4.57 & 74.00 \tiny $\pm$ 6.65 & 94.72 \tiny $\pm$ 4.00 & 85.51 \tiny $\pm$ 3.03 & 74.20 \tiny $\pm$ 2.46 \\ 
        DICE & 27.72 \tiny $\pm$ 4.21 & 40.92 \tiny $\pm$ 2.82 & 59.71 \tiny $\pm$ 0.17 & 81.04 \tiny $\pm$ 4.21 & 92.78 \tiny $\pm$ 0.28 & 60.47 \tiny $\pm$ 4.39 & 83.24 \tiny $\pm$ 2.22 & 87.75 \tiny $\pm$ 2.61 & 87.37 \tiny $\pm$ 1.24 & 70.55 \tiny $\pm$ 1.32 \\ 
        MCDropout & 39.43 \tiny $\pm$ 2.45 & 28.45 \tiny $\pm$ 3.56 & 75.70 \tiny $\pm$ 0.85 & 70.63 \tiny $\pm$ 4.53 & 92.67 \tiny $\pm$ 0.29 & 50.03 \tiny $\pm$ 5.16 & 63.23 \tiny $\pm$ 15.03 & 86.45 \tiny $\pm$ 2.95 & 86.43 \tiny $\pm$ 6.56 & 86.71 \tiny $\pm$ 3.08 \\ 
        Energy & 31.03 \tiny $\pm$ 4.19 & 32.01 \tiny $\pm$ 3.43 & 63.81 \tiny $\pm$ 1.02 & 79.77 \tiny $\pm$ 3.43 & 93.13 \tiny $\pm$ 0.19 & 51.86 \tiny $\pm$ 1.99 & 77.45 \tiny $\pm$ 7.26 & 86.78 \tiny $\pm$ 1.64 & 86.92 \tiny $\pm$ 3.01 & 79.24 \tiny $\pm$ 3.11 \\ 
        fDBD & 29.25 \tiny $\pm$ 1.79 & 18.81 \tiny $\pm$ 1.63 & 71.31 \tiny $\pm$ 1.92 & 37.19 \tiny $\pm$ 3.93 & 95.05 \tiny $\pm$ 0.33 & 16.43 \tiny $\pm$ 6.01 & 11.92 \tiny $\pm$ 3.33 & 56.69 \tiny $\pm$ 10.33 & 26.71 \tiny $\pm$ 4.09 & 96.74 \tiny $\pm$ 1.18 \\
\hline
    \end{tabularx}
    \caption{Far-OoD on DenseNet-201.}
    \label{201-far} 
\end{table*}

\begin{table*}[!htbp]
    \centering
    \scriptsize
    
    \begin{tabularx}{\textwidth}{l >{\Centering\arraybackslash}X >{\Centering\arraybackslash}p{2.5cm} >{\Centering\arraybackslash}X >{\Centering\arraybackslash}p{2.5cm} >{\Centering\arraybackslash}X} 
    \toprule % 顶部粗线
        \textbf{Method} & \textbf{FPR95-ID}↓ & \textbf{FPR95-OoD}↓ & \textbf{FPR99-ID}↓ & \textbf{FPR99-OoD}↓ & \textbf{AUROC}↑ \\ 
        \midrule % 中间细线
        ASH & 46.85 \tiny $\pm$ 4.61 & 37.22 \tiny $\pm$ 2.97 & 83.83 \tiny $\pm$ 11.70 & 61.45 \tiny $\pm$ 2.27 & 91.04 \tiny $\pm$ 0.63 \\ 
        DICE & 22.44 \tiny $\pm$ 3.25 & 31.02 \tiny $\pm$ 7.62 & 60.69 \tiny $\pm$ 5.55 & 79.16 \tiny $\pm$ 11.18 & 94.05 \tiny $\pm$ 0.70 \\ 
        MCDropout & 37.44 \tiny $\pm$ 1.46 & 24.41 \tiny $\pm$ 5.23 & 74.74 \tiny $\pm$ 1.67 & 67.04 \tiny $\pm$ 11.20 & 93.34 \tiny $\pm$ 0.64 \\ 
        Energy & 24.50 \tiny $\pm$ 3.10 & 23.58 \tiny $\pm$ 5.24 & 61.63 \tiny $\pm$ 4.31 & 75.99 \tiny $\pm$ 12.26 & 94.40 \tiny $\pm$ 0.57 \\ 
        fDBD & 30.10 \tiny $\pm$ 1.27 & 19.41 \tiny $\pm$ 2.33 & 69.99 \tiny $\pm$ 3.34 & 33.28 \tiny $\pm$ 0.78 & 95.11 \tiny $\pm$ 0.28 \\ 
        GEN & 25.93 \tiny $\pm$ 1.90 & 18.64 \tiny $\pm$ 3.70 & 64.89 \tiny $\pm$ 4.14 & 52.49 \tiny $\pm$ 9.05 & 95.07 \tiny $\pm$ 0.48 \\ 
        GradNorm & 77.97 \tiny $\pm$ 6.00 & 87.77 \tiny $\pm$ 3.60 & 85.26 \tiny $\pm$ 4.27 & 96.04 \tiny $\pm$ 1.65 & 64.17 \tiny $\pm$ 5.31 \\
\end{tabularx}
    \caption{Near-OoD on DenseNet-201.}
    \label{201-near} 
\end{table*}

\subsection{SE-ResNeXt-50}
\Cref{se-far,se-near} show the comprehensive performance of the SE-ResNeXt-50 network on the Far-OoD and Near-OoD benchmarks.

\begin{table*}[htbp!]
    \centering
    \scriptsize
    \begin{tabularx}{\textwidth}{@{}l *{10}{Y}@{}} % 使用tabularx，总宽度为\textwidth，Y列会自动调整宽度
    \toprule % 顶部粗线
    \multirow{2}{*}{\parbox[c]{1.5cm}{%
        \centering
        \vspace{3ex}% 在“Method”文本的上方增加 0.5ex 的空间
        \textbf{Method}%
    }} & \multicolumn{5}{c}{\textbf{Far-OoD(Bubbles \& Particles)}} & \multicolumn{5}{c}{\textbf{Far-OoD(General)}} \\
    \cmidrule(lr){2-6} \cmidrule(lr){7-11} % 部分横线
    & \textbf{FPR95-ID}↓ & \textbf{FPR95-OoD}↓ & \textbf{FPR99-ID}↓ & \textbf{FPR99-OoD}↓ & \textbf{AUROC}↑ & \textbf{FPR95-ID}↓ & \textbf{FPR95-OoD}↓ & \textbf{FPR99-ID}↓ & \textbf{FPR99-OoD}↓ & \textbf{AUROC}↑ \\
    \midrule % 中间细线
        ASH & 89.39 \tiny $\pm$ 14.21 & 85.11 \tiny $\pm$ 5.08 & 95.68 \tiny $\pm$ 6.04 & 93.68 \tiny $\pm$ 3.11 & 63.32 \tiny $\pm$ 8.53 & 89.30 \tiny $\pm$ 15.12 & 90.04 \tiny $\pm$ 7.58 & 98.10 \tiny $\pm$ 2.69 & 96.50 \tiny $\pm$ 2.90 & 45.13 \tiny $\pm$ 22.91 \\ 
        DICE & 35.57 \tiny $\pm$ 3.77 & 50.73 \tiny $\pm$ 5.21 & 62.76 \tiny $\pm$ 3.97 & 85.02 \tiny $\pm$ 0.06 & 90.22 \tiny $\pm$ 1.08 & 34.80 \tiny $\pm$ 5.91 & 54.80 \tiny $\pm$ 13.08 & 65.70 \tiny $\pm$ 7.32 & 79.37 \tiny $\pm$ 8.42 & 89.68 \tiny $\pm$ 1.94 \\ 
        MCDropout & 46.67 \tiny $\pm$ 2.36 & 40.68 \tiny $\pm$ 6.56 & 73.66 \tiny $\pm$ 2.65 & 75.40 \tiny $\pm$ 5.71 & 90.13 \tiny $\pm$ 1.23 & 59.79 \tiny $\pm$ 13.55 & 44.02 \tiny $\pm$ 12.45 & 85.33 \tiny $\pm$ 7.95 & 75.73 \tiny $\pm$ 11.49 & 86.74 \tiny $\pm$ 4.51 \\ 
        Energy & 36.51 \tiny $\pm$ 3.35 & 42.23 \tiny $\pm$ 7.83 & 66.57 \tiny $\pm$ 0.28 & 85.36 \tiny $\pm$ 0.72 & 91.45 \tiny $\pm$ 1.06 & 43.43 \tiny $\pm$ 16.27 & 45.69 \tiny $\pm$ 7.51 & 78.62 \tiny $\pm$ 7.61 & 78.82 \tiny $\pm$ 7.12 & 90.11 \tiny $\pm$ 2.19 \\ 
        fDBD & 36.64 \tiny $\pm$ 2.87 & 32.95 \tiny $\pm$ 5.81 & 72.82 \tiny $\pm$ 1.55 & 67.94 \tiny $\pm$ 10.20 & 92.26 \tiny $\pm$ 1.17 & 46.48 \tiny $\pm$ 16.85 & 29.89 \tiny $\pm$ 10.52 & 83.05 \tiny $\pm$ 8.40 & 49.48 \tiny $\pm$ 16.42 & 88.61 \tiny $\pm$ 5.17 \\
\hline
    \end{tabularx}
    \caption{Far-OoD on SE-ResNeXt-50.}
    \label{se-far}
\end{table*}

\begin{table*}[!htbp]
    \centering
    \scriptsize
    
    \begin{tabularx}{\textwidth}{l >{\Centering\arraybackslash}X >{\Centering\arraybackslash}p{2.5cm} >{\Centering\arraybackslash}X >{\Centering\arraybackslash}p{2.5cm} >{\Centering\arraybackslash}X} 
    \toprule % 顶部粗线
        \textbf{Method} & \textbf{FPR95-ID}↓ & \textbf{FPR95-OoD}↓ & \textbf{FPR99-ID}↓ & \textbf{FPR99-OoD}↓ & \textbf{AUROC}↑ \\ 
        \midrule % 中间细线
        ASH & 90.91 \tiny $\pm$ 12.00 & 73.13 \tiny $\pm$ 4.03 & 98.28 \tiny $\pm$ 2.37 & 88.15 \tiny $\pm$ 0.98 & 67.02 \tiny $\pm$ 8.17 \\ 
        DICE & 27.94 \tiny $\pm$ 3.33 & 35.80 \tiny $\pm$ 2.75 & 59.41 \tiny $\pm$ 4.87 & 73.66 \tiny $\pm$ 4.89 & 93.19 \tiny $\pm$ 0.56 \\ 
        MCDropout & 43.79 \tiny $\pm$ 2.20 & 26.51 \tiny $\pm$ 1.75 & 71.21 \tiny $\pm$ 4.00 & 56.46 \tiny $\pm$ 3.08 & 92.48 \tiny $\pm$ 0.51 \\ 
        Energy & 28.00 \tiny $\pm$ 1.77 & 23.00 \tiny $\pm$ 3.00 & 63.52 \tiny $\pm$ 2.60 & 63.99 \tiny $\pm$ 4.34 & 94.35 \tiny $\pm$ 0.35 \\ 
        fDBD & 30.48 \tiny $\pm$ 1.27 & 18.95 \tiny $\pm$ 1.52 & 69.87 \tiny $\pm$ 3.11 & 30.68 \tiny $\pm$ 1.34 & 95.02 \tiny $\pm$ 0.24 \\ 
        GEN & 29.57 \tiny $\pm$ 2.94 & 18.20 \tiny $\pm$ 1.82 & 63.61 \tiny $\pm$ 3.79 & 35.74 \tiny $\pm$ 0.90 & 95.15 \tiny $\pm$ 0.40 \\ 
        GradNorm & 99.30 \tiny $\pm$ 0.99 & 92.26 \tiny $\pm$ 1.21 & 99.94 \tiny $\pm$ 0.08 & 97.68 \tiny $\pm$ 0.82 & 49.21 \tiny $\pm$ 4.73 \\
\end{tabularx}
    \caption{Near-OoD on SE-ResNeXt-50.}
    \label{se-near}
\end{table*}

\subsection{ViT}
\Cref{vit-far,vit-near}  show the comprehensive performance of the ViT network on the Far-OoD and Near-OoD benchmarks.

\begin{table*}[htbp!]
    \centering
    \scriptsize
    \begin{tabularx}{\textwidth}{@{}l *{10}{Y}@{}} % 使用tabularx，总宽度为\textwidth，Y列会自动调整宽度
    \toprule % 顶部粗线
    \multirow{2}{*}{\parbox[c]{1.5cm}{%
        \centering
        \vspace{3ex}% 在“Method”文本的上方增加 0.5ex 的空间
        \textbf{Method}%
    }} & \multicolumn{5}{c}{\textbf{Far-OoD(Bubbles \& Particles)}} & \multicolumn{5}{c}{\textbf{Far-OoD(General)}} \\
    \cmidrule(lr){2-6} \cmidrule(lr){7-11} % 部分横线
    & \textbf{FPR95-ID}↓ & \textbf{FPR95-OoD}↓ & \textbf{FPR99-ID}↓ & \textbf{FPR99-OoD}↓ & \textbf{AUROC}↑ & \textbf{FPR95-ID}↓ & \textbf{FPR95-OoD}↓ & \textbf{FPR99-ID}↓ & \textbf{FPR99-OoD}↓ & \textbf{AUROC}↑ \\
    \midrule % 中间细线
        ASH & 93.84 \tiny $\pm$ 1.87 & 94.74 \tiny $\pm$ 3.61 & 97.94 \tiny $\pm$ 1.01 & 98.85 \tiny $\pm$ 0.81 & 51.22 \tiny $\pm$ 5.38 & 99.64 \tiny $\pm$ 0.25 & 72.79 \tiny $\pm$ 3.88 & 99.98 \tiny $\pm$ 0.02 & 84.06 \tiny $\pm$ 1.52 & 58.53 \tiny $\pm$ 1.05 \\ 
        DICE & 68.72 \tiny $\pm$ 4.69 & 54.40 \tiny $\pm$ 4.94 & 90.06 \tiny $\pm$ 2.02 & 71.99 \tiny $\pm$ 5.59 & 82.19 \tiny $\pm$ 2.01 & 84.53 \tiny $\pm$ 10.98 & 44.95 \tiny $\pm$ 11.82 & 97.22 \tiny $\pm$ 2.92 & 55.78 \tiny $\pm$ 10.57 & 76.49 \tiny $\pm$ 11.28 \\ 
        MCDropout & 76.52 \tiny $\pm$ 0.96 & 56.86 \tiny $\pm$ 4.28 & 93.14 \tiny $\pm$ 0.19 & 78.66 \tiny $\pm$ 2.59 & 80.53 \tiny $\pm$ 1.42 & 70.29 \tiny $\pm$ 8.15 & 43.30 \tiny $\pm$ 5.60 & 90.39 \tiny $\pm$ 4.16 & 60.85 \tiny $\pm$ 5.65 & 84.63 \tiny $\pm$ 2.97 \\ 
        Energy & 57.44 \tiny $\pm$ 5.19 & 42.73 \tiny $\pm$ 4.94 & 87.94 \tiny $\pm$ 1.48 & 64.10 \tiny $\pm$ 4.68 & 87.53 \tiny $\pm$ 1.74 & 36.48 \tiny $\pm$ 3.05 & 18.22 \tiny $\pm$ 1.87 & 83.46 \tiny $\pm$ 9.45 & 30.12 \tiny $\pm$ 3.04 & 94.05 \tiny $\pm$ 0.52 \\ 
        fDBD & 49.53 \tiny $\pm$ 4.25 & 33.41 \tiny $\pm$ 4.25 & 82.01 \tiny $\pm$ 1.61 & 53.63 \tiny $\pm$ 5.05 & 90.63 \tiny $\pm$ 1.27 & 31.38 \tiny $\pm$ 12.99 & 14.50 \tiny $\pm$ 3.55 & 76.34 \tiny $\pm$ 7.43 & 24.81 \tiny $\pm$ 4.01 & 95.06 \tiny $\pm$ 1.81 \\
\hline
    \end{tabularx}
    \caption{Far-OoD on ViT.}
    \label{vit-far}
\end{table*}

\begin{table*}[!htbp]
    \centering
    \scriptsize
    
    \begin{tabularx}{\textwidth}{l >{\Centering\arraybackslash}X >{\Centering\arraybackslash}p{2.5cm} >{\Centering\arraybackslash}X >{\Centering\arraybackslash}p{2.5cm} >{\Centering\arraybackslash}X} 
    \toprule % 顶部粗线
        \textbf{Method} & \textbf{FPR95-ID}↓ & \textbf{FPR95-OoD}↓ & \textbf{FPR99-ID}↓ & \textbf{FPR99-OoD}↓ & \textbf{AUROC}↑ \\ 
        \midrule % 中间细线
        ASH & 95.63 \tiny $\pm$ 1.54 & 94.36 \tiny $\pm$ 1.32 & 98.51 \tiny $\pm$ 0.91 & 98.84 \tiny $\pm$ 0.38 & 52.41 \tiny $\pm$ 2.66 \\ 
        DICE & 79.40 \tiny $\pm$ 4.97 & 72.98 \tiny $\pm$ 1.25 & 95.72 \tiny $\pm$ 0.68 & 83.75 \tiny $\pm$ 2.44 & 74.35 \tiny $\pm$ 2.85 \\ 
        MCDropout & 77.16 \tiny $\pm$ 0.86 & 61.11 \tiny $\pm$ 6.32 & 93.30 \tiny $\pm$ 0.29 & 81.73 \tiny $\pm$ 7.33 & 79.78 \tiny $\pm$ 0.49 \\ 
        Energy & 63.40 \tiny $\pm$ 4.01 & 52.34 \tiny $\pm$ 8.65 & 91.81 \tiny $\pm$ 1.45 & 72.17 \tiny $\pm$ 10.20 & 85.81 \tiny $\pm$ 0.98 \\ 
        fDBD & 53.15 \tiny $\pm$ 1.90 & 56.78 \tiny $\pm$ 16.50 & 86.77 \tiny $\pm$ 0.72 & 77.89 \tiny $\pm$ 15.94 & 87.39 \tiny $\pm$ 1.77 \\ 
        GEN & 58.71 \tiny $\pm$ 2.94 & 50.24 \tiny $\pm$ 10.76 & 88.40 \tiny $\pm$ 1.65 & 70.22 \tiny $\pm$ 12.19 & 87.00 \tiny $\pm$ 0.92 \\ 
        GradNorm & 67.72 \tiny $\pm$ 3.63 & 63.24 \tiny $\pm$ 2.75 & 90.33 \tiny $\pm$ 2.44 & 85.43 \tiny $\pm$ 1.28 & 81.05 \tiny $\pm$ 1.96 \\
\end{tabularx}
    \caption{Near-OoD on ViT.}
    \label{vit-near}
\end{table*}

% % 
% Having the supplementary compiled together with the main paper means that:
% % 
% \begin{itemize}
% \item The supplementary can back-reference sections of the main paper, for example, we can refer to \cref{sec:intro};
% \item The main paper can forward reference sub-sections within the supplementary explicitly (e.g. referring to a particular experiment); 
% \item When submitted to arXiv, the supplementary will already included at the end of the paper.
% \end{itemize}
% % 
% To split the supplementary pages from the main paper, you can use \href{https://support.apple.com/en-ca/guide/preview/prvw11793/mac#:/:text=Delete%20a%20page%20from%20a,or%20choose%20Edit%20%3E%20Delete).}{Preview (on macOS)}, \href{https://www.adobe.com/acrobat/how-to/delete-pages-from-pdf.html#:/:text=Choose%20%E2%80%9CTools%E2%80%9D%20%3E%20%E2%80%9COrganize,or%20pages%20from%20the%20file.}{Adobe Acrobat} (on all OSs), as well as \href{https://superuser.com/questions/517986/is-it-possible-to-delete-some-pages-of-a-pdf-document}{command line tools}.

\end{document}